\documentclass[journal]{IEEEtran}
\usepackage{amsmath,amsfonts}
\usepackage{bm}
\usepackage{mathrsfs}
\usepackage{algorithmic}
\usepackage{algorithm}
\usepackage{array}
\usepackage[caption=false,font=normalsize,labelfont=sf,textfont=sf]{subfig}
\usepackage{textcomp}
\usepackage{stfloats}
\usepackage{url}
\usepackage{verbatim}
\usepackage{graphicx}
\usepackage{graphics}
\usepackage{cite}
\usepackage{multirow}

\def\BibTeX{{\rm B\kern-.05em{\sc i\kern-.025em b}\kern-.08em
		T\kern-.1667em\lower.7ex\hbox{E}\kern-.125emX}}
\usepackage{balance}

\def\etal{\emph{et al.}}
\def\ie{\emph{i.e.}}
\def\eg{\emph{e.g.}}

\graphicspath{{Figs/}}

\begin{document}

\title{Self-Paced Deep Regression Forests\\ with Consideration of Ranking Fairness}

\author{Lili Pan$^*$, Mingming Meng, Yazhou Ren, Yali Zheng, Zenglin Xu,~\IEEEmembership{Senior Member,~IEEE}
\thanks{Lili Pan and Mingming Meng are with the School of Information and Communication Engineering, University of Electronic Science and Technology of China, Chengdu, 611731, and Yangtze Delta Region Institute, University of Electronic Science and Technology of China, Quzhou, 32400,  China.}
\thanks{Yazhou Ren is with the School of Computer Science and Engineering, and Yali Zheng is with the School of Automation Engineering, University of Electronic Science and Technology of China, Chengdu, 611731, China.}
\thanks{Zenglin Xu is with the School of Computer Science and Engineering, Harbin Institute of Technology, Shenzhen, and also affiliated with Peng Cheng Lab, Shenzhen, 518055,  China.}}

\markboth{IEEE Transaction on Neural Networks and Learning Systems,~Vol.~15, No.~5, August~2022}%
{Shell \MakeLowercase{\textit{et al.}}: A Sample Article Using IEEEtran.cls for IEEE Journals}


\maketitle

\begin{abstract} 
	Deep discriminative models (DDMs), \eg~deep regression forests and deep decision forests, have been extensively studied recently to solve problems such as facial age estimation, head pose estimation, \emph{etc.}.
	Due to a shortage of well-labeled data that does not have noise and imbalanced distribution problems, learning DDMs is always challenging.
	Existing methods usually tackle these challenges through learning more discriminative features or re-weighting samples.
	We argue that learning DDMs gradually, from easy to hard, is more reasonable, for two reasons. First, this is more consistent with the cognitive process of human beings. Second, noisy as well as underrepresented examples can be distinguished by virtue of previously learned knowledge.
	Thus, we resort to a gradual learning strategy---self-paced learning (SPL).
	Then, a natural question arises: can SPL lead DDMs to achieve more robust and less biased solutions?
	To answer this question, this paper proposes a new SPL method: easy and underrepresented examples first, for learning DDMs.
	This tackles the fundamental ranking and selection problem in SPL from a new perspective: fairness.
	Our idea is fundamental and can be easily combined with a variety of DDMs.
	Extensive experimental results on three computer vision tasks, \ie, facial age estimation, head pose estimation, and gaze estimation, show our new method gains considerable performance improvement in both accuracy and fairness.
	Source code is available at \url{https://github.com/learninginvision/SPU}.
\end{abstract}


\section{Introduction}
\IEEEPARstart{D}{eep} discriminative models (DDMs) have been extensively studied recently to solve a variety of computer vision problems, with applications in image classification, action recognition, and object detection, to name a few.
DDMs use deep neural networks (DNNs)~\cite{chendeepage, Yan2014Age,chen_using_2017,huang2018mixture} to formulate the input-output mapping in discriminative models.
Due to a shortage of well-labeled training data, \ie~without noise and imbalanced distribution problems, learning DDMs is particularly challenging. 
Considerable literature has grown up around the theme of how to learn DDMs effectively~\cite{rothe2018deep,huang_soft-margin_2017,ruiz2018fine}.

One typical approach is to learn more discriminative features through rather deep neural networks and cost-sensitive discriminative functions~\cite{gao_age_2018,parkhi2015deep,niu_ordinal_2016,rothe2018deep,agustsson2017anchored}.
The other typical approaches are to reweight training samples according to estimation errors~\cite{cui2019class,khan2019striking} or gradient directions~\cite{ren2018learning} (\ie~meta learning).
Although these approaches can partially alleviate the noise and imbalanced distribution problem of training data, they are not consistent with the cognitive process of human beings; the difference is we learn new things based on previously learned knowledge.
In addition, these approaches do not provide a  way to distinguish noisy and underrepresented examples, and thus cannot fundamentally avoid noisy and biased solutions.

\begin{figure*}[t]
	\centering
	\includegraphics[width=0.9\textwidth]{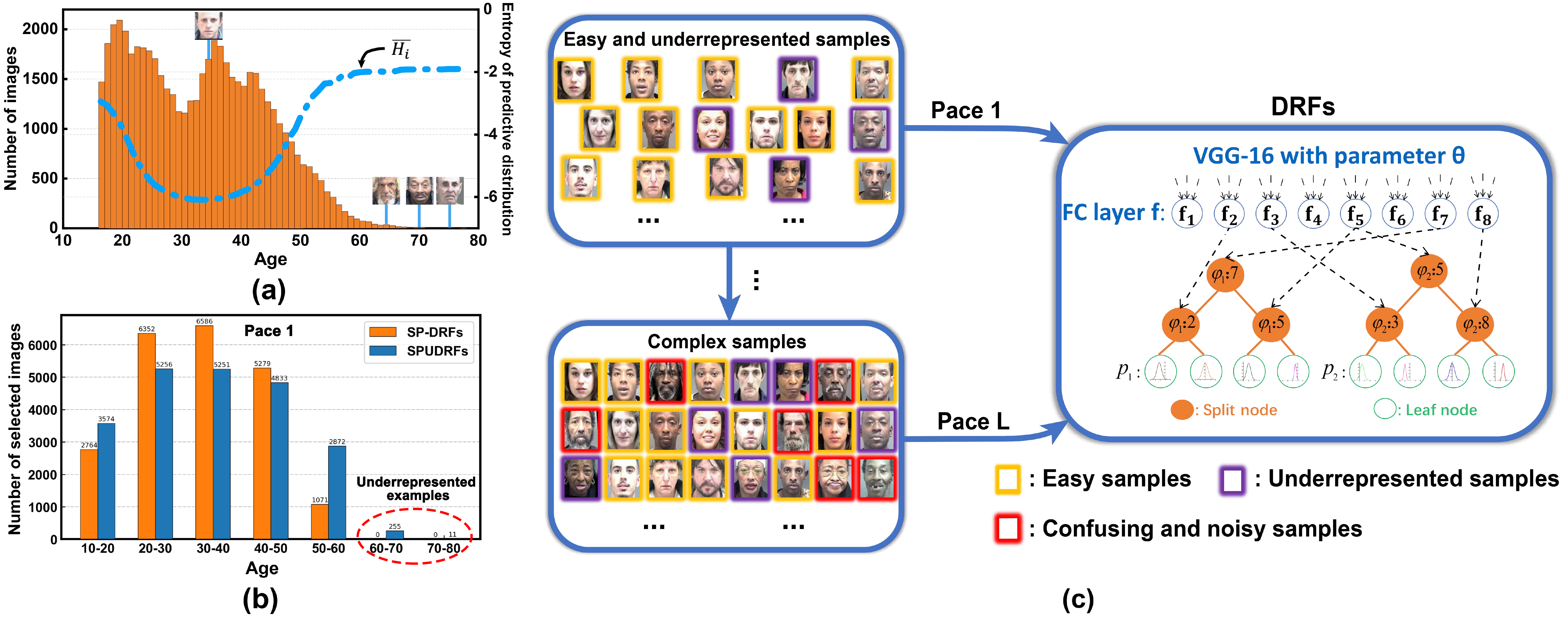}
	\caption{The Motivation of considering underrepresented examples in SPL. \textbf{(a):} The histogram shows the number of facial images of different ages, and the average entropy curve indicates the predictive uncertainty. We observe that the \emph{underrepresented examples} always have high entropy. \textbf{(b):} Underrepresented examples are selected only in SPUDRFS at an early pace. Because the underrepresented examples tend to have relatively large loss, they would not be selected at an early pace. \textbf{(c):} A new self-paced learning method: easy and underrepresented examples first. This builds on the intuition that the underrepresented examples are prone to incur unfair treatment since they are the ``minority'' in training data.}
	\label{Figure1}
\end{figure*}
  Intuitively, learning DDMs from easy to hard may be more reasonable because already-learned knowledge, in such a process, can be leveraged to learn DDMs. 
In addition to this, the noisy and underrepresented examples could be recognized by virtue of already-learned knowledge.
Hence, we resort to a gradual learning strategy  \ie, self-paced learning~\cite{Kumar2010Self, jiang2015self, ma2017self, ijcai2017}.
So far, however, there are few studies on learning DDMs in a self-paced manner.
Then, a natural question arises: \emph{can SPL guide DDMs to achieve more robust and less biased solutions?}

An underlying problem in SPL, which is firstly shown by this study, is that it assumes the distribution of training data is balanced, and thus the bias of solutions may be further exacerbated when such an assumption fails.
The reason is that the underrepresented examples, due to \emph{large} prediction error, would not be selected for training at early paces, thereby incurring unfair treatment.
This means existing SPL methods have fairness issue.

To address the fairness issue in existing SPL methods as well as answer the above questions, this paper proposes a new self-paced learning method for learning DDMs.
It tackles the fundamental ranking and selection problems in SPL from a new perspective: fairness.
On one hand, our new method keeps SPL's advantage in robustness.
On the other hand, it prevents seriously biased solutions induced by SPL.
The major contributions of this work include:
\begin{itemize}
	\item We for the first time, show that existing SPL methods may lead to seriously biased solutions. To address this problem, we propose a new SPL method: easy and underrepresented examples first. This combines with a typical DDM, \ie, deep regression forests (DRFs), can lead to a new model---deep regression forests with consideration of underrepresented examples (SPUDRFs). The new model shows considerable performance improvement in both accuracy and fairness.
	\item To further promote robustness, in SPUDRFs, we define capped likelihood function with respect to DRFs’ parameters so as to further exclude noisy examples.
	\item To measure regression fairness, we define a fairness metric for regression problems, which can reflect fairness concerning some sensitive attributes.  
	\item  For verification, we validate SPUDRFs on three computer vision tasks: (\romannumeral1)  facial age estimation, (\romannumeral2) head pose estimation,  and (\romannumeral3) gaze estimation.
	Extensive experimental results on the Morph \uppercase\expandafter{\romannumeral2}~\cite{ricanek2006morph}, FG-NET~\cite{panis2016overview}, BIWI~\cite{fanelli2013random} , BU3DFE~\cite{pan2016mixture}  and MPIIGaze~\cite{zhang2015appearance} datasets demonstrate the efficacy of our proposed new self-paced method.
	On all the above tasks, SPUDRFs almost achieve state-of-the-art performance.
\end{itemize}

A preliminary version of this work was published in~\cite{pan2020self}.
This paper significantly improves~\cite{pan2020self} in the following aspects. 
(\romannumeral1) We extend our model to incorporate capped likelihood, which could further promote robustness.
The robust SPUDRFs can recognize and exclude examples with labeling noise,  whereas the original is unable to do so, and only places more emphasis on reliable examples. 
(\romannumeral2) We extend our model to combine with various weighting schemes, whereas \cite{pan2020self} merely adopts a mixture weighting scheme.
(\romannumeral3) We define the fairness metric for regression in this work and show obvious fairness improvement of SPUDRFs against the original SP-DRFs. 
(\romannumeral4) We also evaluate SPUDRFs on a new computer vision task, \ie, gaze estimation, and demonstrate its validity on the MPIIGaze dataset.
(\romannumeral5) Both robustness and fairness evaluation results are added, along with more comprehensive discussions.

This paper is organized as follows.  Sec.~\ref{sec:related work} introduces related work on SPL, and DDMs for facial age estimation, head pose estimation and gaze estimation. Sec.~\ref{sec:SPUDRFs} details our proposed SPUDRFs. Sec.~\ref{sec:robust SPUDRFs} introduces robust SPUDRFs.
Sec.~\ref{sec:fairness metric} defines a fairness metric for regression problem.
Sec.~\ref{sec:experiment} exhibits and analyzes the experimental results on three computer vision tasks and five related datasets. Sec.~\ref{sec:discussion} discusses the strengths and potential weaknesses of this work, followed by Sec.~\ref{sec:conclusion} concluding this paper with perspectives.

\section{Related Work}
\label{sec:related work}
This section reviews SPL methods and DDMs-based facial age estimation, head pose estimation and gaze estimation.

\subsection{Self-Paced Learning}
Self-paced learning, as a gradual learning method, builds on the intuition that, rather than considering all training samples simultaneously, the learning should be presented with the training data from easy to difficult~\cite{Kumar2010Self,meng_theoretical_2017}.
To date, a great deal of study on self-paced learning (SPL) has been undertaken, mostly about learning conventional discriminative models, \eg~SVM, logistic regressor.
There also exists some literature in which SPL is employed for clustering problems~\cite{Huang2021DSMVC,Huang2021NSMVC, Guo2019adaptive}.

Learning conventional discriminative models in a self-paced manner exhibits superiority over traditional methods.
In~\cite{Kumar2010Self}, Kumar \etal~proposed the original SPL framework in a regime that suggests processing the samples in order of an easy to hard order. 
In~\cite{jiang2015self}, Zhao \etal~generalized the hard weighting scheme in SPL to a soft weighting scheme, which promoted discrimination accuracy.
In~\cite{ma2017self}, Ma \etal~proposed self-paced co-training, where SPL is applied for multi-view or multi-modality problems.
In~\cite{ren2018self}, Ren \etal~introduced capped-norm into the objective function of SPL, so as to further exclude the interference of noisy examples.
In fact, the above work can be cast as incorporating SPL with conventional classifiers, \eg, SVM, logistic regressors, \emph{etc.}.
To incorporate SPL with regression models, in~\cite{han2017self}, Han \etal~learned a mixture of regressors in a self-paced manner, and added an $\ell_{1,2}$ norm regularizer to avoid poorly conditioned linear sub-regressors.

In computer vision community, recently some researchers have realized that learning DDMs in a self-paced manner may lead to more robust solutions.
One of the few studies is~\cite{ijcai2017}, where Li \etal~sought to enhance the learning robustness of CNNs by SPL.
However, they omitted one important problem in learning discriminative models: the imbalanced distribution  of training data.

Our work is inspired by the existing studies~\cite{yang2019self,jiang2014self} which take the class diversity in sample selection of SPL into consideration.
Jiang \etal~\cite{jiang2014self} proposed to ensure the class diversity of samples at the early paces in self-paced training.
Yang \etal~\cite{yang2019self} defined a metric called ``complexity of image category" to measure the number of samples in each category, as well as jointly classifying difficulty, and used it to select samples.
The two methods mentioned above find that a lack of class diversity in sample selection may result in seriously biased solutions.
However, what causes this problem has not been studied. 
This work shows that the cause is fundamentally the unfairness of sample ranking in SPL, where underrepresented examples may often have large losses and not be selected at early paces.
On the other hand, \cite{yang2019self,jiang2014self} are only suitable for classification, but not for regression in which the output targets are continuous and high dimensional.
This paper will go further in this direction, aiming to tackle the fundamental drawback in SPL---ranking unfairness---and to integrate SPL with DDMs.

\subsection{Facial Age Estimation}
Facial age estimation has been extensively studied for a decade in the computer vision community.
In recent years, a large and growing body of literature~\cite{niu_ordinal_2016,chen_using_2017,gao_age_2018,shen_deep_2018,li2019bridgenet} has been proposed for age estimation with varying degrees of success.
Ordinal-based approaches~\cite{niu_ordinal_2016,chen_using_2017} are the best-known and have demonstrated improved results.
For example, Niu \etal~proposed to estimating age through a set of sequential binary queries---each query refers to a comparison with a predefined age---to exploit the inter-relationship (ordinal information) among age labels.
Furthermore, Gao \etal~\cite{gao_age_2018} proposed DLDL-v2 to explore the underlying age distribution patterns, so as to effectively accommodate age ambiguity.
Shen \etal~\cite{shen_deep_2018} used VGG-16 to extract facial features, and mapped the extracted features to age by deep regression forests (DRFs).
Li \etal~\cite{li2019bridgenet} proposed BridgeNet to predict age by mixing local regression results, where local regressors are learned on partitioned data space. 
Deng \etal~\cite{deng2021pml} proposed an age classification method to deal with long-tailed distribution, where a variational margin is used to minimize the influence of head classes that misleads the prediction of tailed samples. 
Overall, these DDMs-based approaches have improved age estimation performance significantly. However, they plausibly ignore one problem: the interference with facial images duo to PIE (\ie~pose, illumination and expression) variation, occlusion, misalignment and so forth.
More importantly, these approaches mostly ignore the fact that the training data is not distributed uniformally.

\subsection{Head Pose Estimation}
Head pose estimation has attracted vast attention from the computer vision community for a long time.
Recently, an increasing amount of DDMs-based methods have been proposed for head pose estimation and have dramatically boosted estimation accuracy.
Of these, Riegler \etal~\cite{riegler2013hough} utilized convolutional neural networks
(CNNs) to extract patch features within facial images, and more precise results were achieved.
In~\cite{huang2018mixture}, Huang \etal~proposed multi-modal deep regression networks, with multi-layer perceptron (MLP) architecture, to fuse the RGB and depth images for head pose estimation.
In~\cite{wang2019deep}, Wang \etal~proposed a deep coarse-to-fine network to further boost pose estimation accuracy.
In~\cite{ruiz2018fine}, Ruiz \etal~used a large, synthetically expanded head pose dataset to train a rather deep multi-loss convolutional neural networks (CNNs), gaining obvious accuracy improvement.
In~\cite{kuhnke2019deep}, Kuhnke \etal~proposed a domain adaptation method for head pose estimation where shared label spaces across different domains are assumed.
Although accuracy has improved, handling noisy examples in the above models is not straightforward. 
Moreover, the methods mentioned above seldom deal with imbalanced training data, which may widely exist in visual problems.

\subsection{Gaze Estimation}
Appearance-based gaze estimation learns a mapping from facial or eye image to gaze.
Recently, a considerable amount of DDMs based approaches have been proposed for person-independent gaze estimation.
For example,  in~\cite{zhang2017s}, Zhang \etal~used spatial weights VGG-16 to encode face image to gaze direction. 
The spatial weights were used to flexibly suppress or enhance features from different facial regions. 
In the literature~\cite{krafka2016eye}, Krafka \etal~implemented a CNNs-based gaze tracker by mapping captured images from the left eye, right eye and face to gaze. The location and size of the head in the original image are provided for mapping networks. 
In~\cite{fischer2018rt}, Fischer \etal~adopted two backbone networks to extract features from two eye regions for gaze regression, and achieved improved robustness. 
Instead of predicting gaze directly, in~\cite{park2018deep}, Park \etal~proposed another method which maps the single eye image to an intermediate pictorial representation as additional supervision to simplify 3D gaze direction estimation. 
Wang \etal~\cite{wang2019generalizing} proposed a Bayesian adversarial learning approach to address the appearance, head pose variations in gaze estimation, and overfitting problem.
Xiong \etal~\cite{xiong2019mixed} proposed mixed effect neural networks to combine global and person specific gaze estimation results together.
Meanwhile, other studies show that facial images can help gaze estimation.
In \cite{biswas2021appearance, cheng2020gaze, cheng2020coarse}, multi-channel estimation architectures were adopted to utilize facial image and eye images jointly. 
These methods outperform the ones that only use eye images as input for gaze estimation.
In addition to person-independent methods, there is some literature about person-specific gaze adaption methods~\cite{park2019few, yu2019improving, chen2020offset}, which have boosted performance dramatically when a few annotated examples are used for calibration.
These methods indeed improve gaze estimation results; however, whether we should learn DDMs for gaze estimation in a gradual learning manner has not been discussed, nor has there been discussion of how to deal with imbalanced distribution of training data.

\section{Self-Paced Deep Regression Forests with Consideration on Underrepresented Samples}
\label{sec:SPUDRFs}
This section first reviews the basic concepts in DRFs; and introduces the objective formulation for SPUDRFs, as well as variant weighting strategies and an underrepresented example augmentation method; and finally details the optimization algorithm.

\subsection{Preliminaries}

Deep regression forests (DRFs), as a deep regression model, connect deep neural netwoks (DNNs) to regression forests.
We start by reviewing the basic concepts in DRFs~\cite{shen_deep_2018}.

\noindent \textbf{Deep Regression Tree.} DRFs usually consist of a number of deep regression trees, each of which, given input-output pairs $\left\{\mathbf{x}_i, y_i\right\}_{i=1}^N$, map extracted features through DNNs to target output by passing a regression tree. Further, a regression tree $\mathcal{T}$ consists of split (or decision) nodes $\mathcal{N}$ and leaf (or prediction) nodes $\mathcal{L}$~\cite{shen_deep_2018} (see Fig.~\ref{Figure1}).
Specifically, each split node $n \in \mathcal{N}$ determines whether a sample $\mathbf{x}_i$ goes to left or right sub-tree; each leaf node $\ell \in \mathcal{L}$ represents a Gaussian distribution $p_{\ell}(y_i)$ with mean $\mu_l$ and variance $\sigma^2_l$---parameters of output distribution defined for each tree $\mathcal{T}$.

\noindent \textbf{Split Node.}
Split node represents a split function $s_{n}(\mathbf{x}_i ; \bm{\Theta}) : \mathbf{x}_i \rightarrow[0,1]$, where $\bm{\Theta}$ denotes the parameters of DNNs, as in Fig.~\ref{Figure1}(c).
Conventionally, $s_{n}(\mathbf{x}_i ; \bm{\Theta})$ is formulated as $\sigma\left(\mathbf{f}_{\varphi(n)}(\mathbf{x}_i; \bm{\Theta})\right)$, where $\sigma(\cdot)$ denotes a sigmoid function while $\varphi(\cdot)$ denotes an index function specifying the $\varphi(n)$-th element of $\mathbf{f}(\mathbf{x}_i; \bm{\Theta})$ in correspondence with the split node $n$, and $\mathbf{f}(\mathbf{x}_i; \bm{\Theta})$ denotes the extracted features through DNNs.
An example to illustrate the sketch chart of the DRFs is shown in Fig.~\ref{Figure1}(c), where $\varphi_1$ and $\varphi_2$ are two index functions for two trees.
Finally, the probability of the sample $\mathbf{x}_i$ falling into the leaf node $\ell$ is given by:
\begin{equation}
	\label{Eq.1}
	\omega_\ell( \mathbf{x}_i | \bm{\Theta)}=\prod_{n \in \mathcal{N}} s_{n}(\mathbf{x}_i ; \bm{\Theta})^{[\ell \in \mathcal{L}_{n_{\text{left}}}]}\left(1-s_{n}(\mathbf{x}_i ; \bm{\Theta})\right)^{\left[\ell \in \mathcal{L}_{n_{\text{right}}}\right]},
\end{equation}
where $[\cdot]$ denotes an indicator function conditioned on the argument. 
In addition, $\mathcal{L}_{n_\text{left}}$ and $\mathcal{L}_{n_\text{right}}$ correspond to the sets of leaf nodes owned by the sub-trees $\mathcal{T}_{n_\text{left}}$ and $\mathcal{T}_{n_\text{right}}$ rooted at the left and right children ${n}_{l}$ and ${n}_{r}$ of node $n$.

\noindent \textbf{Leaf Node.} 
For a tree $\mathcal{T}$, each leaf node $\ell \in \mathcal{L}$ defines a Gaussian distribution over output $y_i$ conditioned on $\left\{\mu_l,\sigma^2_l\right\}$.
Since each example $\mathbf{x}_i$ has the probability $\omega_\ell(\mathbf{x}_i | \bm{\Theta)}$ to reach leaf node $\ell$, considering all leaf nodes, the predictive distribution over $y_i$ shall sum all leaf distribution weighted by $\omega_\ell(\mathbf{x}_i | \bm{\Theta)}$:
\begin{equation}
	\label{Eq.2}
	p_{\mathcal{T}}(y_i | \mathbf{x}_i ; \bm{\Theta}, \bm{\pi})=\sum_{\ell \in \mathcal{L}} \omega_\ell( \mathbf{x}_i | \bm{\Theta)} p_{\ell}(y_i),
\end{equation}
where $\bm{\pi}$ represents the distribution parameters of all leaf nodes associated with tree $\mathcal{T}$.
Note that $\bm{\pi}$ varies along with tree $\mathcal{T}_k$, and thus we rewrite them in terms of $\bm{\pi}_k$.

\noindent \textbf{Forests of Regression Trees.}
Since a forest $\mathcal{F}$ consists of a number of deep regression trees $\left\{\mathcal{T}_1,...,\mathcal{T}_k\right\}$, the predictive output distribution shall consider all trees:
\begin{equation}
	\label{Eq.3}
	p_{\mathcal{F}}\left(y_i|\mathbf{x}_i,\bm{\Theta},\bm{\Pi} \right)
	=
	\frac{1}{K}\sum_{k=1}^K p_{\mathcal{T}_k}\left(y_i|\mathbf{x}_i; \bm{\Theta}, \bm{\pi}_k\right),
\end{equation}
where $K$ is the number of trees and $\bm{\Pi}=\left\{\bm{\pi}_1,...,\bm{\pi}_K\right\}$.
$p_{\mathcal{F}}\left(y_i|\mathbf{x}_i,\bm{\Theta},\bm{\Pi} \right)$ denotes the likelihood that the $i^{th}$ sample has output $y_i$.

\subsection{Objective}
\label{Uncertainty}
\begin{figure}
	\centering
	\includegraphics[width=0.50\textwidth]{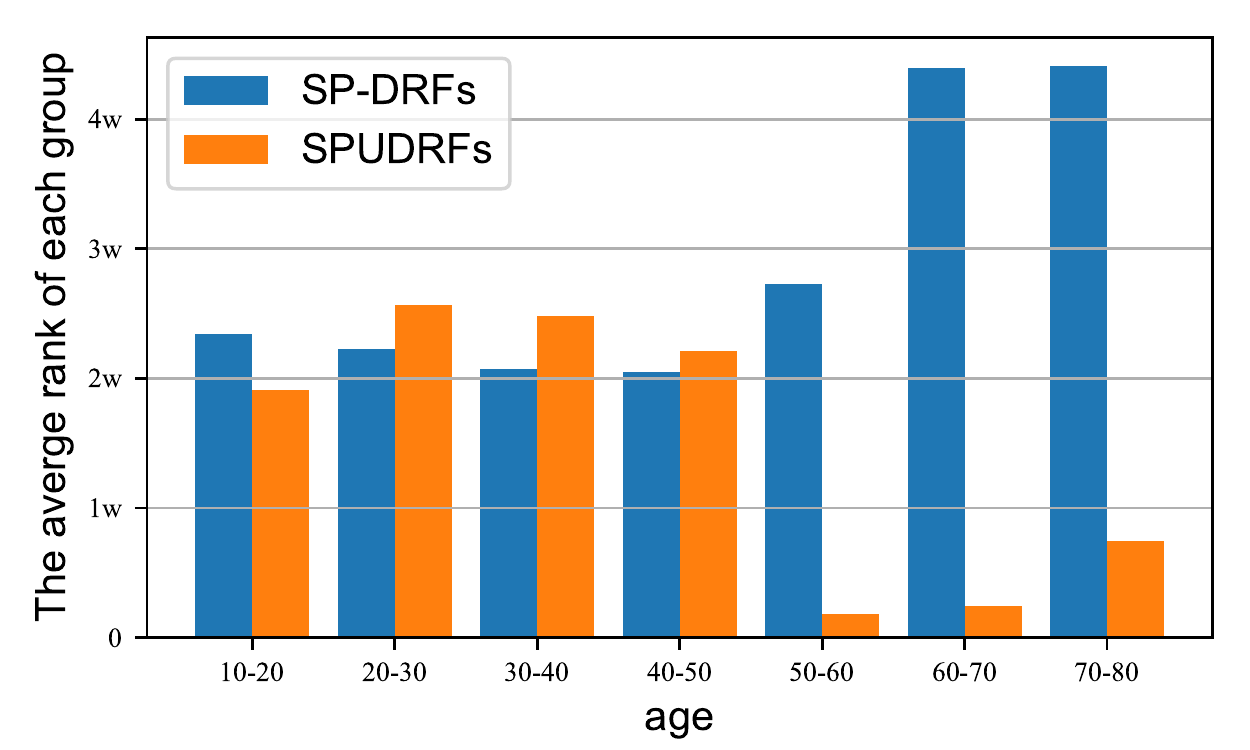}
	\caption{The average rank of each age group in the $1^{st}$ pace. SP-DRFs tend to rank the underrepresented examples at the end. In contrast, SPUDRFs tend to rank the underrepresented examples in the front to ensure they are selected for training from the very beginning.}
	\label{grouprank}
\end{figure}

\noindent\textbf{Underrepresented Examples.} Considering underrepresented examples in SPL is one of the main contributions of this work.
In this work, underrepresented examples mean ``minority''.
The underrepresented level could be measured by predictive uncertainty (\ie~entropy).
In fact, underrepresented examples may incur unfair treatment in early paces due to imbalanced data distribution.
Our proposed method tackles the ranking and selection problems in SPL from a new perspective: \emph{fairness}.

\noindent\textbf{Entropy.} Given a sample $\mathbf{x}_i$, its predictive uncertainty is formulated by calculating the entropy of the conditional output distribution $p_{\mathcal{F}}\left(y_i|\mathbf{x}_i,\bm{\Theta},\bm{\Pi} \right)$:
\begin{equation}
	\label{Eq.4}
	H\left [p_{\mathcal{F}}\left(y_i|\mathbf{x}_i,\bm{\Theta},\bm{\Pi} \right)\right] = \frac{1}{K}\sum^K_{k=1}H\left [p_{\mathcal{T}_k}\left(y_i|\mathbf{x}_i,\bm{\Theta}, \bm{\pi}_k \right)\right],
\end{equation}
where $H\left[ \cdot\right ]$ denotes entropy, and the entropy which corresponds to the $k^{th}$ tree is:
\begin{multline}
	\label{Eq.5}
	H\left [p_{\mathcal{T}_k}\left(y_i|\mathbf{x}_i,\bm{\Theta}, \bm{\pi}_k \right)\right] =\\ -\int p_{\mathcal{T}_k}\left(y_i|\mathbf{x}_i,\bm{\Theta}, \bm{\pi}_k \right)\ln p_{\mathcal{T}_k}\left(y_i|\mathbf{x}_i,\bm{\Theta}, \bm{\pi}_k \right) dy_i.
\end{multline}
Given $\mathbf{x}_i$, the larger the entropy $H\left [p_{\mathcal{F}}\left(y_i|\mathbf{x}_i,\bm{\Theta},\bm{\Pi} \right)\right]$ is, the more uncertain the prediction should be, \ie, the more underrepresented the sample $\mathbf{x}_i$ is.

The predictive distribution $p_{\mathcal{T}_k}\left(y_i|\mathbf{x}_i; \bm{\Theta}, \bm{\pi}_k\right)$ takes the form $\sum_{\ell \in \mathcal{L}} \omega_\ell( \mathbf{x}_i | \bm{\Theta)} p_{\ell}(y_i)$.
Because the integral of mixture of Gaussians in Eq.~(\ref{Eq.5}) is non-trivial, we resort to Monte Carlo sampling.
However, because this has a large computational cost~\cite{huber2008entropy}, we turn to use the lower bound of this integral to approximate the integration:
\begin{equation}
	\label{Eq.6}
	H\left [p_{\mathcal{T}_k}\left(y_i|\mathbf{x}_i,\bm{\Theta}, \bm{\pi}_k \right)\right]\approx\frac{1}{2}\sum_{\ell\in\mathcal{L}}\omega_\ell(\mathbf{x}_i|\mathbf{\Theta})\left[\ln \left(2\pi \sigma_\ell^2\right)+1\right].
\end{equation}

\noindent\textbf{Formulation.} Let $v_i$ denote a latent variable indicating whether the $i^{th}$ sample is selected $(v_i = 1)$ or not $(v_i = 0)$ for training.
Our objective is to jointly maximize the log likelihood function with
respect to DRFs’ parameters $\bm{\Theta}$ and $\bm{\Pi}$, and learn the latent variables $\mathbf{v}=\left(v_1,...,v_N\right)^T$.
Fig.~\ref{grouprank} shows that the original self-paced method may miss the underrepresented examples at an early pace, resulting in ignorance.
Considering fairness, we prefer to select the underrepresented examples, which probably have higher predictive uncertainty (\ie~entropy), particularly at an early pace.
Therefore, we maximize the likelihood function of DRFs coupled with the sample selection term meanwhile considering ranking fairness,
\begin{equation}
	\label{Eq.7}
	\max_{\bm{\Theta},\bm{\Pi}, \mathbf{v}} \sum_{i=1}^{N} v_{i} \left \{ \log p_{\mathcal{F}}\left(y_i|\mathbf{x}_i,\bm{\Theta},\bm{\Pi} \right) + \gamma H_i \right \}  + \lambda\sum_{i=1}^N v_i ,
\end{equation}
where $\lambda$ controls learning pace ($\lambda\geq0$), $\gamma$ is a trade-off between loss and uncertainty ($\lambda\geq0$), $H_i$ denotes the entropy of output distribution $p_{\mathcal{F}}\left(y_i|\mathbf{x}_i,\bm{\Theta},\bm{\Pi} \right)$.
When $\gamma$ decays to $0$, the objective function is equivalent to the log-likelihood function with respect to DRFs’ parameters $\bm{\Theta}$ and $\bm{\Pi}$.
In Eq.~(\ref{Eq.7}), each sample is weighted by $v_i$, and whether $\log p_{\mathcal{F}}\left(y_i|\mathbf{x}_i,\bm{\Theta},\bm{\Pi} \right) + \gamma H_i>-\lambda$ determines
the $i^{th}$ sample is selected.
That is, the sample with high likelihood value or entropy may be selected.
The optimal $v_i^*$ is:
\begin{align}
	\label{Eq.8}
	v_i^* = \left\{ \begin{array}{ll}
		1 & \textrm{if $\log p_{\mathcal{F}i} + \gamma H_i > -\lambda$}\\
		0 & \textrm{otherwise}
	\end{array} \right.,
\end{align}
where $p_{\mathcal{F}}\left(y_i|\mathbf{x}_i,\bm{\Theta},\bm{\Pi} \right)$ is denoted by $ p_{\mathcal{F}i}$ for simplicity, and we use this representation in all following parts.
Iteratively increasing $\lambda$ and decreasing $\gamma$, samples are dynamically involved for training DRFs, starting with easy and underrepresented examples and ending up with all samples.
Thus, SPUDRFs are prone to achieve more robust and less biased solutions.

One might argue that the noisy and hard examples tend to have high predictive uncertainty, with the result that they are selected in the early pace.
In fact, from Eq.~(\ref{Eq.8}), we observe that whether a sample is selected is determined by both its predictive uncertainty and the log likelihood of being predicted correctly.
The noisy and hard examples probably have relatively large loss \ie~low log likelihood $\log p_{\mathcal{F}i}$, and would not be selected in the early pace.

\subsection{Weighting Schemes}
Weighting each sample according to its importance would be more reasonable in SPUDRFs.
In the previous section, we adopted a hard weighting scheme in SPUDRFs, as defined in Eq.~(\ref{Eq.7}), where one sample selected or not is determined by $v_i$.
Such a weighting scheme appears to omit the importance of samples.
SPUDRFs can easily incorporate other weighting schemes, including mixture weighting and soft weighting~\cite{jiang2014easy}.

\noindent\textbf{Mixture Weighting.}
Mixture weighting scheme~\cite{jiang2014easy} weights selected sample by its importance, \ie, $0\leq v_i \leq 1$.
The objective function under mixture weighting scheme is:
\begin{equation}
	\label{Eq.9}
	\max_{\bm{\Theta},\bm{\Pi}, \mathbf{v}} \sum_{i=1}^{N} v_{i} \left \{ \log p_{\mathcal{F}i} + \gamma H_i\right \}  + \zeta \sum_{i=1}^N \log\left(v_i + \zeta/\lambda\right) ,
\end{equation}
where $\zeta$ is a parameter controlling the learning pace.
We set $\zeta=\left(\frac{1}{\lambda'}-\frac{1}{\lambda}\right)^{-1}$, and $\lambda>\lambda'>0$ to construct a reasonable soft weighting formulation.
The self-paced regularizer in Eq.~(\ref{Eq.9}) is convex with respect to $v\in\left[0,1\right]$.
Then, setting the partial gradient of Eq.~(\ref{Eq.9}) with respect to $v_i$ to be zero will lead the following equation:
\begin{equation}
	\log p_{\mathcal{F}i} + \gamma H_i + \frac{\zeta}{v_i + \zeta/\lambda} = 0.
\end{equation}
The optimal solution of $v_i$ is given by:
\begin{align}
	v_i^* = \left\{ \begin{array}{ll}
		1 & \textrm{if $\log p_{\mathcal{F}i} + \gamma H_i \geq -\lambda'  $}\\
		0 & \textrm{if $\log p_{\mathcal{F}i} + \gamma H_i \leq -\lambda $}\\
		\frac{-\zeta}{\log p_{\mathcal{F}i} + \gamma H_i} - \zeta/\lambda & \textrm{otherwise}
	\end{array} \right.
	\label{Eq.11}
\end{align}
If either the log likelihood or the entropy is too large,  $v^*_i$ equals 1.
In addition, if the likelihood and entropy are both too small, $v^*_i$ equals 0.
Except for the above two cases, the soft weighting, \ie, the last line of Eq.~(\ref{Eq.11}), is adopted.

\noindent\textbf{Soft Weighting.}
A soft weighting scheme~\cite{jiang2014easy} weights a selected sample with respect to its output likelihood and entropy. Such a weighting scheme includes: linear weighting scheme and logarithmic weighting scheme.

\noindent\textbf{Linear weighting.} This scheme linearly assigns weights to samples with respect to output likelihood and entropy. The objective of SPUDRFs under linear weighting scheme is formulated as:
\begin{equation}
	\max_{\bm{\Theta},\bm{\Pi}, \mathbf{v}} \sum_{i=1}^{N} v_{i} \left \{ \log p_{\mathcal{F}i} + \gamma H_i\right \} - \frac{1}{2} \lambda \sum_{i=1}^N \left(v_{i}^2-2v_{i}\right),
	\label{Eq.12}
\end{equation}
where $\lambda>0$, $v_i\in[0,1]$. We set the partial gradient of Eq.~(\ref{Eq.12}) with respect to $v_{i}$ to be zero, then the optimal solution for $v_i$ is:
\begin{align}
	v_i^* = \left\{ \begin{array}{ll}
		\frac{\log p_{\mathcal{F}i} + \gamma H_i + \lambda}{\lambda} & \textrm{if $\log p_{\mathcal{F}i} + \gamma H_i \geq -\lambda  $}\\
		0 & \textrm{otherwise}
	\end{array} .\right.
	\label{Eq.13}
\end{align}
The larger the log likelihood and entropy are, the higher the weight associated with the $i^{th}$ sample should be. 

\noindent\textbf{Logarithmic weighting.} This scheme penalizes the output likelihood and entropy logarithmically. The objective of SPUDRFs under a logarithmic weighting scheme can be formulated as:
\begin{equation}
	\max_{\bm{\Theta},\bm{\Pi}, \mathbf{v}} \sum_{i=1}^{N} v_{i} \left \{ \log p_{\mathcal{F}i} + \gamma H_i\right \} - \sum_{i=1}^N \left( \zeta v_{i} - \frac{\zeta^{v_i}} {\log \zeta} \right),
	\label{Eq.14}
\end{equation}
where $\zeta=1-\lambda$, $0 < \lambda < 1$ and $v_i\in[0,1]$. Similarly, the optimal solution for $v_i$ is
\begin{align}
	v_i^* = \left\{ \begin{array}{ll}
		\frac{\log \left( \zeta - \log p_{\mathcal{F}i} - \gamma H_i \right)}{\log \zeta} & \textrm{if $\log p_{\mathcal{F}i} + \gamma H_i \geq -\lambda  $}\\
		0 & \textrm{otherwise}
	\end{array} .\right.
	\label{Eq.15}
\end{align}

\subsection{Underrepresented Example Augmentation}

As previously explained, the SPUDRFs method places more emphasis on underrepresented examples and may achieve less biased solutions.
Since the intrinsic reason for SPL's biased solutions is the ignorance of underrepresented examples, we further rebalance training data via distribution reconstruction.
Specifically, we distinguish the underrepresented examples whose $H_i$ are larger than $\beta$, from regular examples at each pace and augment them.
As the number of underrepresented examples increases through augmentation, the label distribution imbalance problem is alleviated.

\subsection{Optimization}
\label{Learning}
To optimize SPUDRFs, as defined in Eq.~(\ref{Eq.7}), we propose a two-step alternative search strategy (ASS) algorithm: (\romannumeral1) For sample selection, update $\mathbf{v}$ with fixed $\bm{\Theta}$ and $\bm{\Pi}$ (\romannumeral2) update $\bm{\Theta}$ and $\bm{\Pi}$ with current fixed sample weights $\mathbf{v}$.
In addition, with fixed $\mathbf{v}$,  our DRFs are learned by alternatively updating $\bm{\Theta}$ and $\bm{\Pi}$.
In \cite{shen_deep_2018}, the parameters $\bm{\Theta}$ for split nodes (\ie~parameters for VGG) are updated through gradient descent since the loss is differentiable with respect to $\bm{\Theta}$.
In comparison, the parameters $\bm{\Pi}$ for leaf nodes are updated by virtue of variational bounding~\cite{shen_deep_2018} when fixing $\bm{\Theta}$.

\renewcommand{\algorithmicrequire}{ \textbf{Input:}}
\renewcommand{\algorithmicensure}{ \textbf{Output:}}
\begin{algorithm}[t]
	\caption{The training process of SPUDRFs.}
	\label{alg:The}
	\begin{algorithmic}[1]
		\REQUIRE
		$\mathcal{D}=\left\{\mathbf{x}_i, \mathbf{y}_i\right\}_{i=1}^N$.
		\ENSURE
		Model parameters $\bm{\Pi}$, $\bm{\Theta}$.
		\STATE Initialize $\bm{\Pi}^{0}$, $\bm{\Theta}^{0}$, $\lambda^0$, $\gamma^0$.
		\STATE \textbf{while} not converged \textbf{do}
		\STATE \quad Update $\mathbf{v}$ by Eq.~(\ref{Eq.8}).
		\STATE \quad \quad \textbf{while} not converged \textbf{do}
		\STATE \quad \quad \quad \quad Randomly select a batch from $\mathcal{D}$.
		\STATE \quad \quad \quad \quad Calculate $H_i$ for each sample by Eq.~(\ref{Eq.6}).
		\STATE \quad \quad \quad \quad Update $\bm{\Theta}$ and $\bm{\Pi}$ to maximize Eq.~(\ref{Eq.7}).
		\STATE \quad \quad \textbf{end while}
		\STATE \quad \quad Increase $\lambda$ and decrease $\gamma$.
		\STATE \textbf{end while}
	\end{algorithmic}
	\label{alg1}
\end{algorithm}

\section{Robust Self-Paced Deep Regression Forests with Consideration on underrepresented examples}
\label{sec:robust SPUDRFs}
As has already been discussed in Sec.~\ref{sec:SPUDRFs}, SPL tends to place more emphasis on reliable examples to achieve more robust solutions.
However, SPL's intrinsic selection scheme can not exclude noisy examples, especially the examples with labeling noise.
To alleviate this drawback, we introduce capped-likelihood function in SPUDRFs, which can render an output likelihood with an especially small value as zero:
\begin{equation}
	\text{cap}(p_{\mathcal{F}i},\epsilon) = \frac{\max(p_{\mathcal{F}i}-\epsilon, 0)}{p_{\mathcal{F}i}-\epsilon}p_{\mathcal{F}i}, 
	\label{cap_equation}
\end{equation}
where $\epsilon$ denotes the threshold and $\epsilon>0$. 
Given the output likelihood $p_{\mathcal{F}_i}$, the capped likelihood $p^c_{\mathcal{F}_i}$ is defined as $\text{cap}\left(p_{\mathcal{F}_i},\epsilon\right)$. 
Because the log-likelihood values of the noisy examples are prone to be
small, such an operation sets capped likelihood $p^c_{\mathcal{F}_i}$ to be negatively infinite. 
Thus, noisy examples would not be selected, especially not examples with labeling noise.
SPUDRFs with capped likelihood are defined as robust SPUDRFs. 
\begin{equation}
	\max_{\bm{\Theta},\bm{\Pi}, \mathbf{v}} \sum_{i=1}^{N} v_{i} \left \{ \log p_{\mathcal{F}i}^{c} + \gamma H_i \right \}  + \lambda\sum_{i=1}^N v_i,
	\label{Eq.17}
\end{equation}
The robust SPUDRFs can be optimized using the optimization method proposed in Sec.~\ref{Learning}. 
Note that, if $p_{\mathcal{F}_i}\leq\epsilon$, the capped likelihood $p^c_{\mathcal{F}_i}=0$, that is, $\log p^c_{\mathcal{F}_i}=-\infty$. 
Since $\lambda$ is a positive factor, maximizing Eq.~(\ref{Eq.17}) must yield $v_i=0$, meaning that the $i^{th}$ sample would not be selected.
By adjusting $\epsilon$, we can exclude a certain ratio of noisy examples to obtain more robust solutions.

\section{Fairness Metric}
\label{sec:fairness metric}
In addition to \emph{accuracy}, fairness should also be an essential metric to measure the performance of a regression model, particularly when the regression targets are related to sensitive attributes.
For example, in age estimation, we expect our model to treat the younger and the elder fairly, \ie, not to result in large MAEs for the former but small MAEs for the latter.
The notion of fairness was originally defined concerning a protected attribute such as gender, race or age.
However, the term fairness has a range of potential definitions. Here, we adopt a notion of fairness that is for sensitive features~\cite{fitzsimons2019general}.
\emph{Defining a fairness metric for regression is one of the contributions of this work.}

The present studies~\cite{agarwal2019fair, berk2021fairness, komiyama2018nonconvex, zafar2017fairness} mostly refer to fairness constraints in regressions, for example, statistical parity or bounded group loss~\cite{agarwal2019fair, komiyama2018nonconvex}, but seldom refers to fairness metrics.
In this work, we define a new fairness metric for regression models.
To be specific, the test dataset is divided into $K$ subsets, each of which is denoted by $\mathbf{D}_k=\left\{\mathbf{x}_i, y_i| y_i \in \mathcal{G}_k\right\}$, and $\mathcal{G}_k$ denotes the $k^{th}$ group.
A fair model is expected to have the same performance on all subsets, which can be described mathematically as:
\begin{equation}
	\label{Eq1_fairness}
	\mathbb{E}_{k}\left[L \left(\hat{y}, y\right)\right]= \mathbb{E}_{l}\left[L\left(\hat{y}, y\right)\right] \quad \forall k,l \in \left\{1,2,..., K\right\}, k\neq l.
\end{equation}
where $\mathbb{E}_k\left[\cdot\right]$ denotes the expectation with respect to the loss $L\left( \cdot \right)$ over group $\mathbf{D}_k$, $\hat{y}$ is the predicted value of the model and $y$ is the real target value. Motivated by $p\%-$rule \cite{zafar2017fairness}, which measure classification fairness, we evaluate fairness between two subsets $\mathbf{D}_k$ and $\mathbf{D}_l$ as follows:
\begin{equation}
	\label{Eq2_fairness}
	f\left( \mathbf{D}_k,\mathbf{D} _l\right) = \min\left(\frac{\mathbb{E}_{k}\left[L \left(\hat{y}, y\right)\right]}{\mathbb{E}_{l}\left[L\left(\hat{y}, y\right)\right]} ,  \frac{\mathbb{E}_{l}\left[L \left(\hat{y}, y\right)\right]}{\mathbb{E}_{k}\left[L \left(\hat{y}, y\right)\right]} \right).
\end{equation}
The loss expectation ratio characterizes the model's fairness on every two subsets. A small ratio means the performance on such two subsets is significantly distinct, reflecting model bias against a particular subset. 
On the contrary, a large ratio indicates the losses on $\mathbf{D}_k$ and $\mathbf{D}_l$ are similar, indicating the model is relatively fair for $\mathbf{D}_k$ and $\mathbf{D}_l$. In particularly, when the ratio is equal to 1, the model satisfies Eq.~(\ref{Eq1_fairness}), and $\mathbf{D}_k$ and $\mathbf{D}_l$ are treated fairly.

\begin{figure}
	\centering
	\subfloat[]{
		\includegraphics[width=0.48\textwidth]{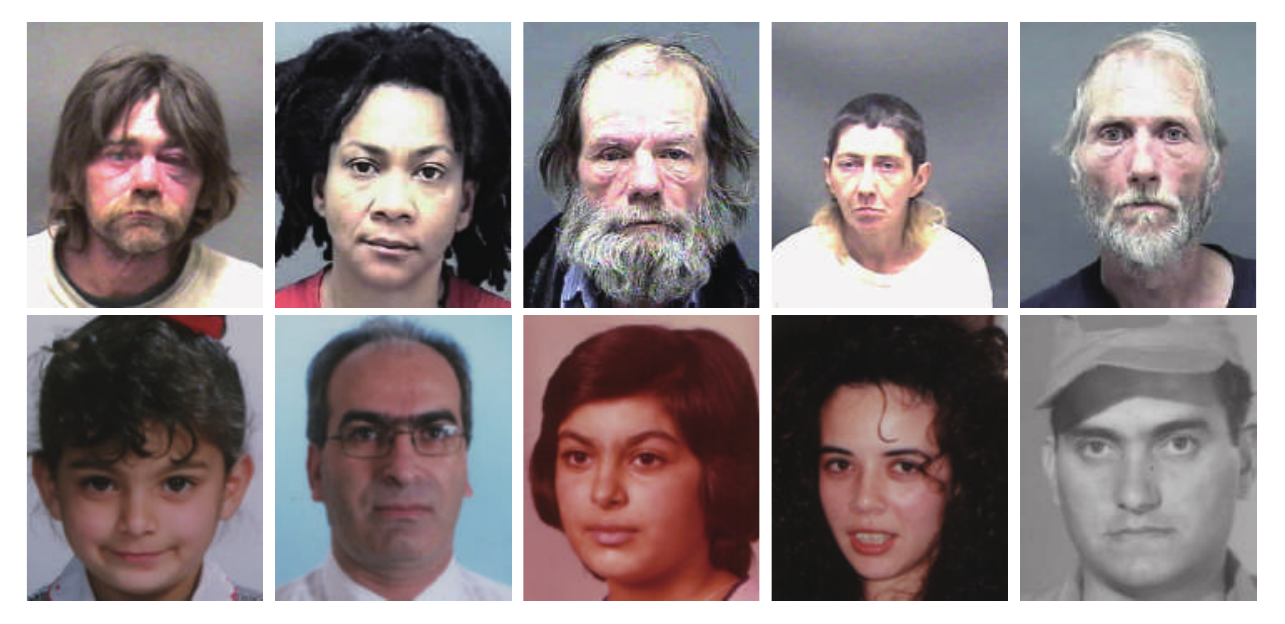}%
		\label{fig:age_samples}}\\
	\subfloat[]{
		\includegraphics[width=0.48\textwidth]{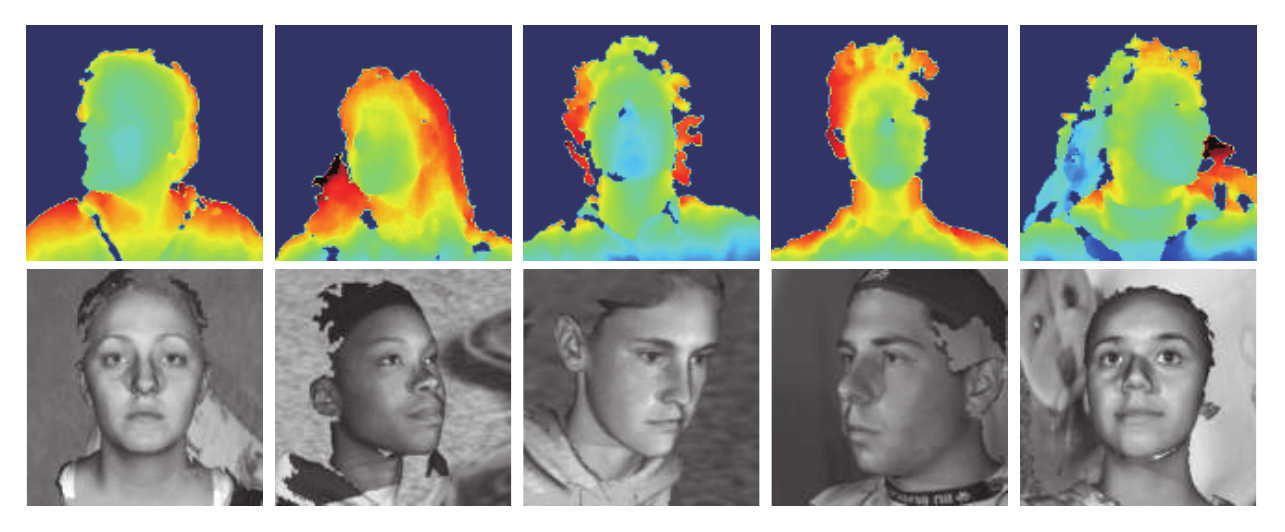}%
		\label{fig:head_samples}}\\
	\subfloat[]{
		\includegraphics[width=0.48\textwidth]{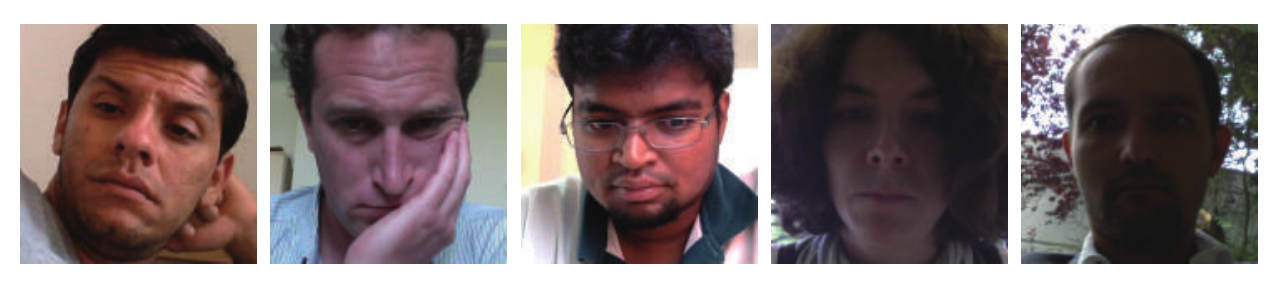}%
		\label{fig:gaze_samples}}\\
	\caption{Example images in the Morph \uppercase\expandafter{\romannumeral2}, FG-NET, BIWI, BU3DFE and MPIIGaze datasets. \textbf{(a):} The images from the Morph \uppercase\expandafter{\romannumeral2} and FG-NET dataset. \textbf{(b):} The images from the BIWI and BU3DFE dataset. \textbf{(c):} The images from MPIIGaze dataset.}
	\label{dataset_samples}
\end{figure}

Finally, the regression fairness is defined as the expectation of the fairness between any two subsets,
\begin{equation}
	\label{Eq3_fairness}
	\text{FAIR} = \mathbb{E}\left[f\left(\mathbf{D}_i,\mathbf{D}_j\right)\right].
\end{equation}
In SPUDRFs, when we set $\gamma \textgreater 0$, all samples are sorted by both likelihood and entropy. As a result, easy and underrepresented samples are selected first, which means samples from all subsets would be selected at early paces. Therefore, our model can alleviate the model's prejudice against different subsets and improve regression fairness. More extensive evaluation results can be found in Sec.~\ref{fairness_section}.

\begin{figure*}[t]
	\centering
	\subfloat[Morph \uppercase\expandafter{\romannumeral2}]{
		\includegraphics[width=1.0\textwidth]{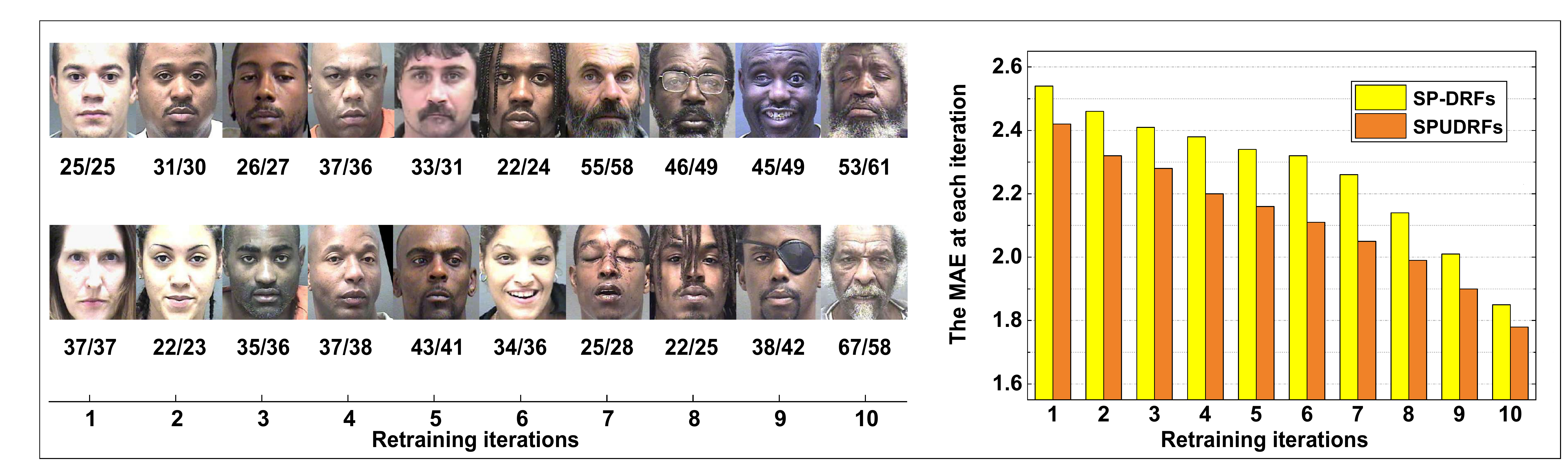}%
		\label{fig:spu_validation_morph}}\\
	\subfloat[MPIIGaze]{
		\includegraphics[width=1.0\textwidth]{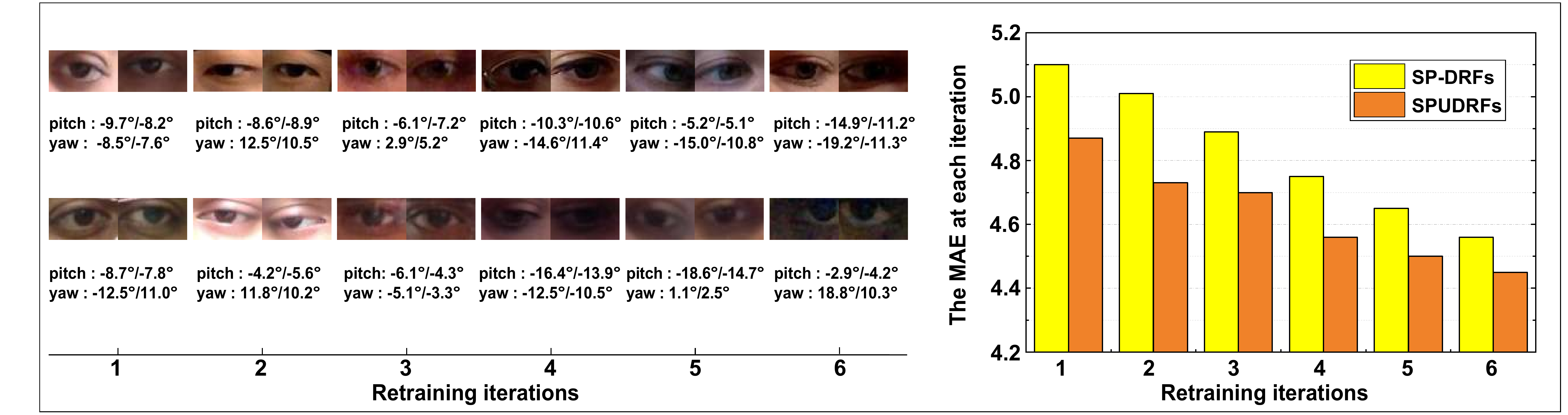}%
		\label{fig:spu_validation_mpii}}\\
	\caption{Visualization of the SPL process on the Morph \uppercase\expandafter{\romannumeral2} and MPIIGaze datasets. \textbf{Left:} The typical worst cases at each pace. The images in each panel are sorted by the increasing paces. The two numbers below each image are the labels (left) and predicted targets (right). \textbf{Right:} MAE comparison between SP-DRFs and SPUDRFs at each pace. The MAE bins are sorted by the increasing paces.}
	\label{SPUDRFs_validation}
\end{figure*}

	\section{Experiment}
\label{sec:experiment} 
\subsection{Datasets and Experimental Setup}
\label{setup}
\noindent\textbf{Dataset.} There are five datasets used in our experiments, namely Morph \uppercase\expandafter{\romannumeral2}, FG-NET, BIWI, BU3DFE and MPIIGaze.

\noindent\textbf{Morph \uppercase\expandafter{\romannumeral2}.} 	The Morph \uppercase\expandafter{\romannumeral2}~\cite{ricanek2006morph} dataset contains $55,13$4 face images of $13618$ individuals with unbalanced gender and ethnicity distributions.
These images are near-frontal pose, neutral expression, and uniform illumination.

\noindent\textbf{FG-NET.} The FG-NET~\cite{panis2016overview} dataset includes $1,002$ color and grey images of $82$ people, with each subject having more than $10$ photos at different ages. 
Since all images were taken in an uncontrolled environment, there is a significant deviation on the lighting, pose, and expression (\ie~PIE) of faces inside the dataset.

\noindent\textbf{BIWI.} The BIWI dataset~\cite{fanelli2013random} includes $15678$ images collected by a Kinect sensor device for different persons, and head poses with pitch, yaw, and roll angles mainly ranging within $\pm 60^{\circ}$, $\pm 75^{\circ}$ and $\pm 50^{\circ}$.
These images are from $20$ subjects, including ten males and six females, where four males have been captured twice with/without glasses.

\noindent\textbf{BU3DFE.} The BU-3DFE dataset \cite{pan2016mixture} is collected from $100$ subjects, of whom $44$ are male, and $56$ are female. Following the work \cite{pan2016mixture}, we randomly rotated the 3D face models to produce $6000$ images with pitch and yaw angles ranging within $\pm 30^{\circ}$ and $\pm 90^{\circ}$.

\noindent\textbf{MPIIGaze.} The MPIIGaze dataset~\cite{zhang2015appearance} includes $213659$ images from $15$ persons. The number of images for each person is between $1498$ and $34745$. The normalized gaze angle is in the range of $[-20^\circ, +5.0^\circ]$ degrees in vertical and $[-25^\circ, +25^\circ]$ degrees in horizontal.
Due to massive deviation of image number amongst different persons, similar to \cite{zhang2015appearance}, $1500$ images from the left and right eyes of each person were chosen for final experiment.

\noindent\textbf{Reprocessing and Data Augmentation.}
MTCNN~\cite{zhang_joint_2016} was used for joint face detection and alignment on the Morph \uppercase\expandafter{\romannumeral2} and FG-NET datasets.
In addition, following~\cite{shen_deep_2018}, two methods were adopted for data augmentation (\romannumeral1) random cropping; and (\romannumeral2) random horizontal flipping. 
On the BIWI and BU3DFE datasets, only random cropping was adopted for augmentation.  
On the MPIIGaze dataset, two normalized eye images (\ie~left and right) were obtained following the work~\cite{zhang2015appearance}, and
only random cropping was used for data augmentation.
\begin{figure*}[t]
	\centering
	\includegraphics[width=0.92\textwidth]{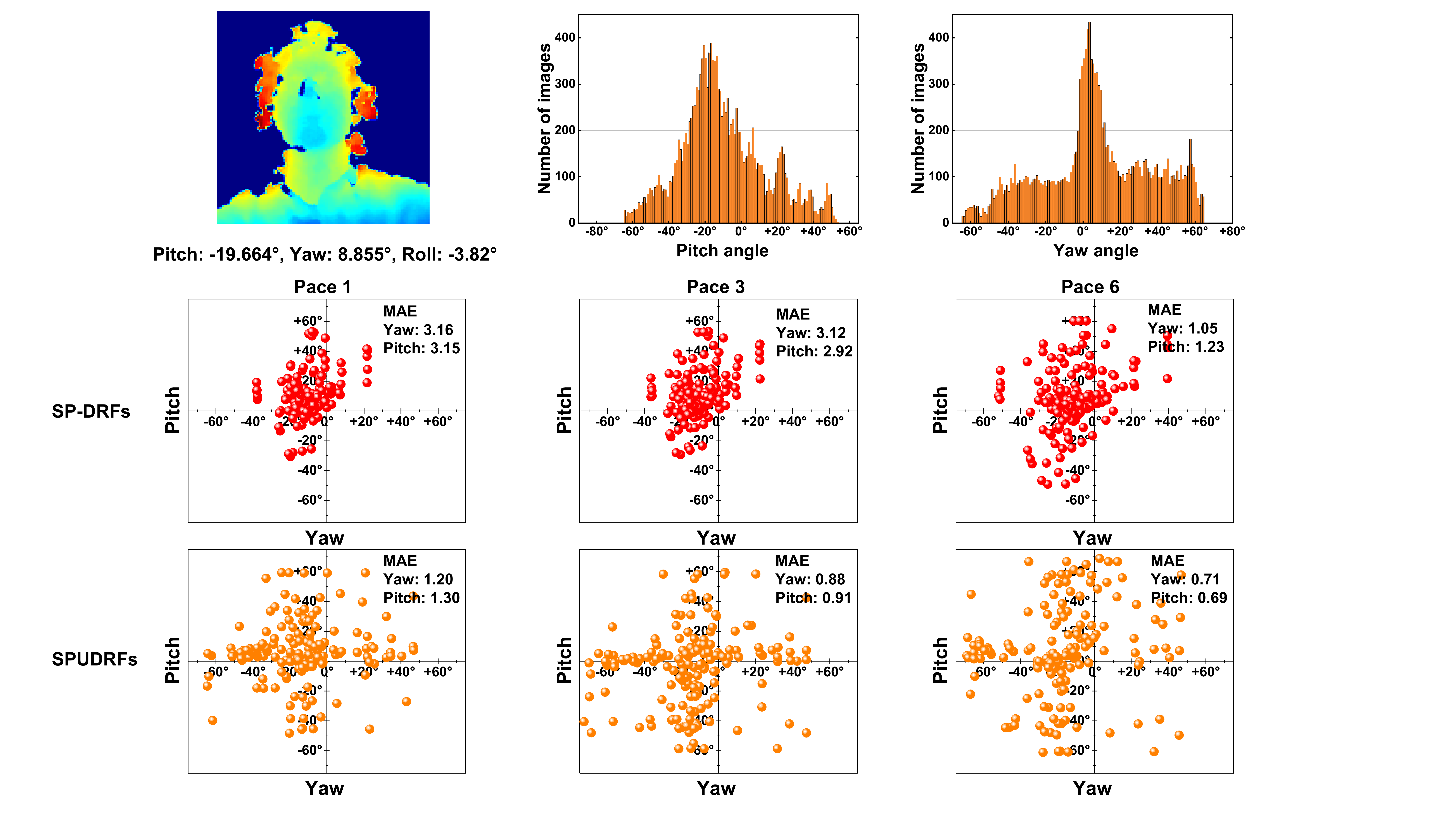}
	\caption{Visualization of leaf node distribution. The $1^{st}$, $3^{rd}$, and $6^{th}$ paces are chosen for visualization. For SP-DRFs, the Gaussian means of leaf nodes (denoted by red points in the second row) concentrate in a small range, resulting in seriously biased solutions. For SPUDRFs, the Gaussian means of leaf nodes (denoted by orange points in the third row) distribute in a reasonable range, resulting in lower MAEs.}
	\label{Uncertainty_efficacy}
\end{figure*}

\noindent\textbf{Parameter Setting.}
VGG-16~\cite{Simonyan2015} was employed as the backbone network of SPUDRFs. 
For MPIIGaze, following \cite{fischer2018rt}, two VGG-16 networks were used.
In addition, the pre-trained models were VGG-16 for MPIIGaze, and VGG-Face~\cite{parkhi2015deep} for other datasets.
When training VGG-16, the batch size were $32$ for Morph~\uppercase\expandafter{\romannumeral2}, BIWI and BU3DFE, $8$ for FG-NET, and $128$ for MPIIGaze.
The drop out ratio was $0.5$.
The maximun iterations in each pace was $80k$ for Morph~\uppercase\expandafter{\romannumeral2}, $10k$ for FG-NET, $20k$ for BIWI and BU3DFE, and $5k$ for MPIIGaze.
SGD optimizer was used for BIWI and Adam optimizer was used for other datasets.
The initial learning rate was $2 \times 10^{-5}$ for Morph~\uppercase\expandafter{\romannumeral2} and BU3DFE, $1 \times 10^{-5}$ for FG-NET, $0.2$ for BIWI, $3 \times 10^{-5}$ for MPIIGaze.
We reduced the learning rate ($×0.5$) per $10k$ iterations for Morph~\uppercase\expandafter{\romannumeral2}, BIWI and BU3DFE.
Here, some hyper-parameter settings are slightly different from the Caffe version~\cite{pan2020self}.	

The hyper-parameters of DRFs were: tree number ($5$), tree depth ($6$), output unit number of feature learning ($128$).
The iterations to update leaf node predictions was $20$, and the number of images to update leaf node predictions was the whole training set for MPIIGaze and $50$ times of batch size for other datasets.
In the first pace, $50\%$ samples were selected for training.
Here, $\lambda$ was set to guarantee the first $50\%$ samples with large $\log p_{\mathcal{F}i} + \gamma H_i$ values were involved.
$\lambda'$ was set to ensure that $10\%\sim20\%$ selected samples adopt soft weighting under a mixture weighing scheme.
To promote efficiency, the samples selected in the previous pace were preserved and the rest were ranked for current sample selection.
In addition, $\gamma$ was initialized to be $15$ for Morph \uppercase\expandafter{\romannumeral2} and BIWI, and $5$ for FG-NET, BU-3DFE, and MPIIGaze.
$\beta$ was used to recognize underrepresented examples that need to be augmented twice in each pace ($1000$ for BIWI, $400$ for BU3DFE, $2000$ for MPIIGaze).
The number of paces was empirically set to be 10 for Morph \uppercase\expandafter{\romannumeral2}, $6$ for BIWI and MPIIGaze, $3$ for FG-NET and BU-3DFE. 
Except for the first pace, an equal proportion of the rest data was gradually involved in each pace.

\noindent\textbf{Evaluation Details.} To evaluate regression accuracy, we used the mean absolute error (MAE).
MAE is defined as $\sum_{i=1}^{N}\left|\hat{y}_{i}-y_{i}\right|/N$, where $\hat{y_{i}}$ represents the predicted output for the $i^{th}$ sample, and $N$ is the total number of images for testing.
In addition, for age estimation, CS calculates the percentage of images sorted in the range of $\left[y_{i}-L, y_{i}+L\right]$: $CS(L)=\sum_{i=1}^{N}\left[\vert\hat{y}_{i}-y_{i}\vert \leq L\right]/N \cdot 100 \%$, where $[ \cdot ]$ denotes an indicator function and $L$ is the error level. 
Further, to measure regression fairness, we adopt our proposed fairness metric FAIR given in Sec.\ref{sec:fairness metric}.

To calculate the above metrics, for Morph \uppercase\expandafter{\romannumeral2}, BIWI and BU3DFE, each dataset was divided into a training set ($80\%$) and a testing dataset ($20\%$). The random division was repeated $5$ times, and the reported MAEs were averaged over $5$ times. 
The leave-one-person-out protocol was used for FG-NET~\cite{shen_deep_2018} and MPIIGaze~\cite{zhang2015appearance}, where one subject was used for testing and the left subjects for training.

\subsection{Validity of SP-DRFs and SPUDRFs}
\label{valid}
This section validates the original SPL method as well as the new proposed SPL method for learning DDMs.
In the following, SP-DRFs denotes self-paced deep regression forests, which learn DRFs using the original SPL method.

\noindent \textbf{Self-Paced Learning.}
Recall that learning DDMs in a gradual learning manner is more consistent with the cognitive process of human beings, and the noisy examples can be distinguished by virtue of learned knowledge.
That means the learning can place more emphasis on ``reliable" examples.
To show this, Fig.~\ref{SPUDRFs_validation}(a) illustrates the representative face images at each learning pace of SP-DRFs on the Morph \uppercase\expandafter{\romannumeral2} dataset, sorted by increasing $\lambda$ and decreasing $\gamma$.
The two numbers below each image are the actual age (left) and predicted age (right).
We observe that the training images involved in each pace become more confusing and noisy step by step, compared to images in early paces.
Since the current model is initialized by the results obtained at the last pace, the updated model is adaptively calibrated by ``reliable'' examples rather than by confusing and noisy ones.
Thus, SP-DRFs have improved MAE and CS compared to DRFs.
We observe that SP-DRFs gain improvement  by 0.33 in MAE $\left(2.17\rightarrow1.84\right)$, by $1.55\%$ in CS $\left(91.3\% \rightarrow92.85\%\right)$, as shown in Fig.~\ref{morph_experiment}(a).

Some representative eye images in the gradual learning sequence for MPIIGaze dataset are shown in Fig.~\ref{SPUDRFs_validation}(b). 
The images in each panel are sorted by increasing $\lambda$ and decreasing $\gamma$. 
The two numbers below each pair of images are the actual gaze direction (left) and the predicted gaze direction (right). 
The same phenomena---the easy examples are prone to be selected in the early paces, while the confusing and noisy ones are prone to be selected in the later paces---can be observed. 
Since the updated model at each pace is adaptively calibrated by ``reliable'' examples rather than by confusing and noisy ones, thus, the MAE associated with each pace decreases step by step.
Finally, the MAE of SP-DRFs decreases to be $4.57$, whereas the MAE of DRFs is $4.62$ (see Table.~\ref{MPII_table}).

\noindent\textbf{Considering Ranking Fairness.}
\label{ExpUnderSamples}
As was mentioned in Sec.~\ref{sec:SPUDRFs}, the existing SPL methods may exacerbate the bias of solutioins.
Fig.~\ref{Uncertainty_efficacy} visualizes the leaf node distributions of SP-DRFs in the gradual learning process.
The Gaussian means $\mu_l$ associated with the $160$ leaf nodes, where each $32$ leaf nodes are defined for $1$ tree, are plotted in each sub-figures.
Three paces, \ie~the $1^{st}$, $3^{rd}$, and $6^{th}$ pace, are randomly chosen for visualization.
For clarity, only pitch and yaw angles are shown.
Meanwhile, the leaf node distributions of SPUDRFs are also visualized in Fig.~\ref{Uncertainty_efficacy}.

In Fig.~\ref{Uncertainty_efficacy}, the comparison between SP-DRFs and SPUDRFs validates our proposed new SPL method.
In SP-DRFs, because the ranking fairness is not considered, the leaf nodes (red points in the $2^{rd}$ row) are only concentrated in a small range that most samples are distributed over, thus leading to seriously biased solutions.
The poor MAEs of SP-DRFs can serve as evidence for this.
In contrast, because the ranking fairness is considered in SPUDRFs, the leaf nodes are distributed over a wide range that could cover underrepresented examples, thus improving performance.
SPUDRFs, in the pitch and yaw directions, achieve the best performance with MAEs of $0.71$ and $0.69$, compared to SP-DRFs with MAEs of $1.05$ and $1.23$ ($47.9\%$ and $78.3\%$ improvements).

\subsection{Comparison with State-of-the-art Methods}
\label{sec:Comparison}

We compare SPUDRFs with other state-of-the-art methods on three vision tasks: age estimation, head pose estimation and gaze estimation.

\begin{figure}[t] 
	\centering 
	\subfloat[]{
		\begin{tabular}{@{}l|c|c}
			\hline
			Method & MAE$\downarrow$ & CS$\uparrow$\\
			\hline
			\hline
			LSVR \cite{guo_human_2009}     & 4.31 & 66.2\% \\
			RCCA \cite{Huerta2014Facial}   & 4.25 & 71.2\% \\
			OHRank \cite{Chang2011Ordinal} & 3.82 & N/A \\
			OR-CNN \cite{niu_ordinal_2016} & 3.27 & 73.0\% \\
			Ranking-CNN \cite{chen_using_2017} & 2.96 & 85.0\% \\
			DLDL-v2 \cite{gao_age_2018}& 1.97 & N/A \\
			DRFs \cite{shen_deep_2018} & 2.17 & 91.3\% \\
			PML \cite{deng2021pml} & 2.15 & N/A \\
			SP-DRFs & 1.84 & 92.85\% \\
			SPUDRFs & \textbf{1.78} & \textbf{93.34\%} \\
			\hline
		\end{tabular}
		\label{tab:morph_mae}
	}\\
	\subfloat[]{
		\includegraphics[width=0.42\textwidth]{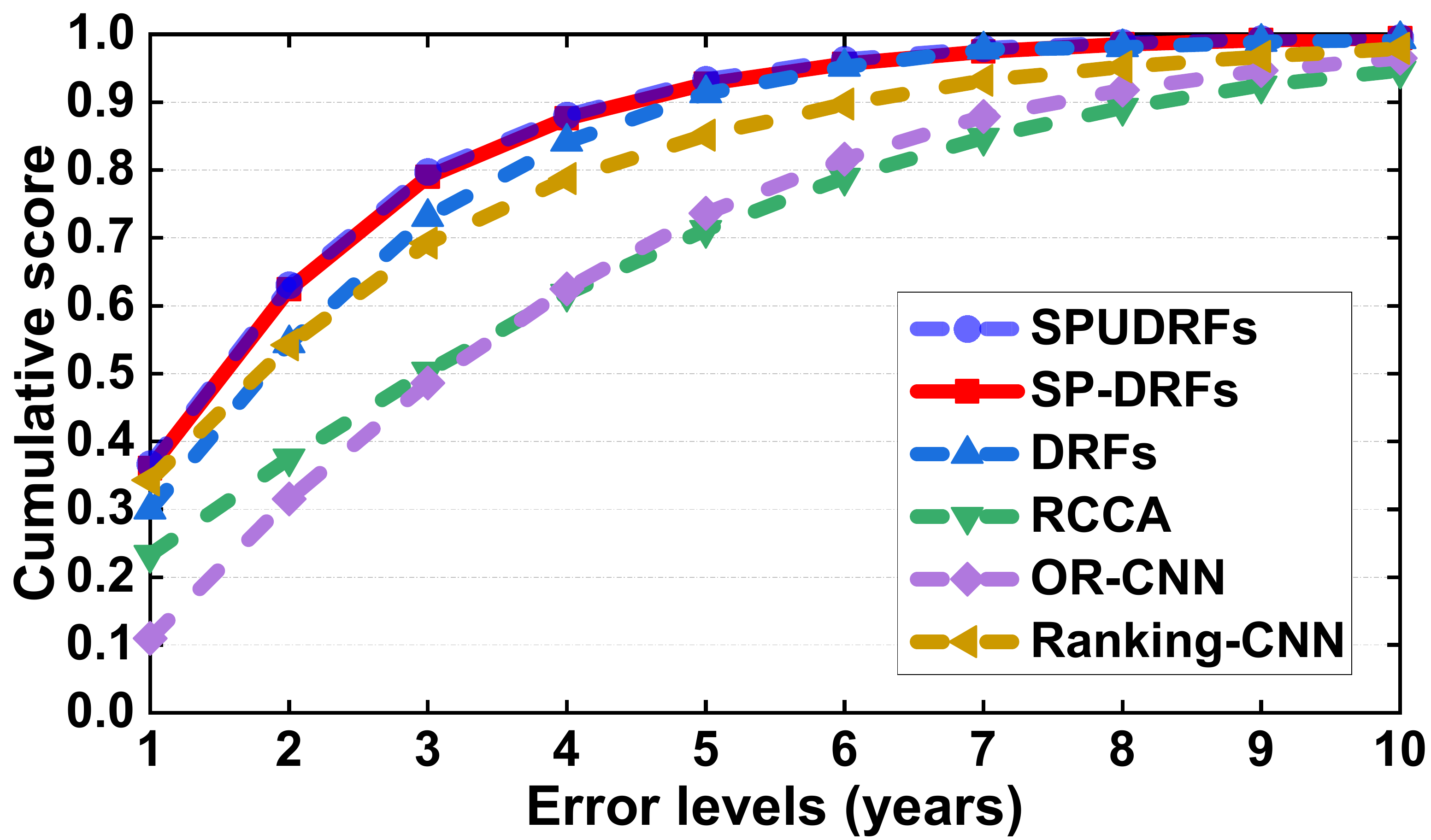}
		\label{fig:morph_cs}
	}                                                  
	\caption{Quantitative comparison with state-of-the-art methods on the Morph \uppercase\expandafter{\romannumeral2} dataset. \textbf{Upper:} MAE comparison results. \textbf{Lower:} CS comparison results.}
	\label{morph_experiment}
\end{figure}

\begin{figure} 
	\centering 
	\subfloat[]{
		\begin{tabular}{@{}l|c|c}
			\hline
			Method & MAE$\downarrow$ & CS$\uparrow$\\
			\hline
			\hline
			IIS-LDL \cite{xin_geng_facial_2013} & 5.77 & N/A \\
			LARR \cite{guodong_guo_image-based_2008} & 5.07 & 68.9\% \\
			MTWGP \cite{Yu2010Multi} & 4.83 & 72.3\% \\
			DIF \cite{han_demographic_2015} & 4.80 & 74.3\% \\
			OHRank \cite{Chang2011Ordinal} & 4.48 & 74.4\% \\
			CAM \cite{Luu2013Contourlet} & 4.12 & 73.5\% \\
			DRFs \cite{shen_deep_2018} & 2.80 & 84.50\% \\
			SP-DRFs& 2.68 & 87.41\% \\
			SPUDRFs & \textbf{2.64} & \textbf{88.59\%}\\
			\hline
		\end{tabular}
		\label{tab:fgnet_mae}
	}\\
	\subfloat[]{
			\includegraphics[width=0.42\textwidth]{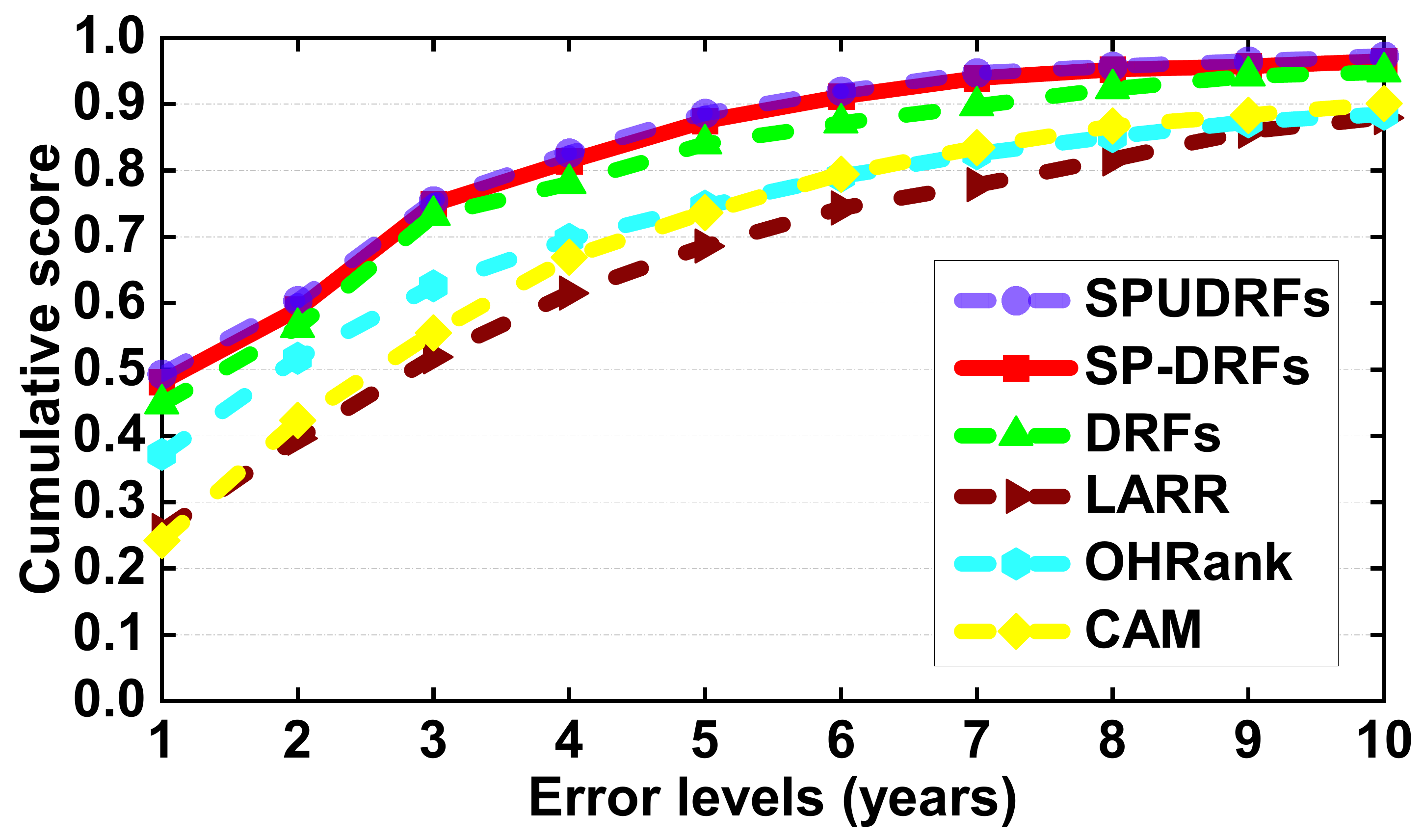}
		\label{fig:fgnet_cs}
	}
	\caption{Quantitative comparison with state-of-the-art methods on the FG-NET dataset. \textbf{Upper:} MAE comparison results. \textbf{Lower:} CS comparison results.}
	\label{fgnet_experiment}
\end{figure}

\noindent\textbf{Results on Age Estimation.} The comparison between our method and other baselines on the Morph \uppercase\expandafter{\romannumeral2} and FG-NET datasets are shown in Fig.~\ref{morph_experiment}(a) and Fig.~\ref{fgnet_experiment}(b).
The baseline methods include: LSVR~\cite{guo_human_2009}, RCCA \cite{Huerta2014Facial}, OHRank~\cite{Chang2011Ordinal}, OR-CNN \cite{niu_ordinal_2016}, Ranking-CNN \cite{chen_using_2017}, DRFs~\cite{shen_deep_2018}, DLDL-v2~\cite{gao_age_2018}, and PML~\cite{deng2021pml}.
The results show some consistent trends.
First, SPUDRFs have superior performance compared to conventional discriminative models, such as LSVR~\cite{guo_human_2009} and OHRank~\cite{Chang2011Ordinal}.
Second, SP-DRFs consistently outperform other DDMs.
Compared to DRFs, our gains in MAE are $0.33$ on Morph \uppercase\expandafter{\romannumeral2} and 0.12 on FG-NET, and in CS are $1.55\%$ and $2.91\%$, respectively.
The promotions demonstrate that learning DRFs in a self-paced manner is more reasonable.
Third, SPUDRFs outperform SP-DRFs on both MAE and CS.

Fig.~\ref{morph_experiment}(b) shows the CS curves of SPUDRFs, SP-DRFs and other baseline methods on the Morph \uppercase\expandafter{\romannumeral2} dataset, and Fig.~\ref{fgnet_experiment}(b) shows the CS curves of different methods on the FG-NET dataset. 
On both datasets above, SPUDRFs consistently outperform other DDMs. 
We observe, the CS of SPUDRFs reaches $93.34\%$ at error level $L=5$, significantly outperforming DRFs by $2.04\%$ on the Morph \uppercase\expandafter{\romannumeral2} dataset. 
We also observe that SPUDRFs outperform DRFs by $4.09\%$ in CS, on the FG-NET dataset.
The CS increase clearly validates our proposed self-paced learning method.

\noindent\textbf{Results on Head Pose Estimation.} 
Recall that considering underrepresented examples in SPL is particularly crucial  when the training data has an imbalanced distribution problem.
Sec.~\ref{valid} has shown the effectiveness of SP-DRFs and SPUDRFs for head pose estimation tasks on the BIWI dataset.
Fig.~\ref{Uncertainty_efficacy} shows the Gaussian means of learned leaf nodes of SP-DRFs and SPUDRFs on the BIWI dataset.
In SP-DRFs, because the underrepresented examples are neglected, the learned leaf nodes are distributed over a small range, resulting in seriously biased solutions.
In SPUDRFs, owing to consideration of underrepresented examples, the learned leaf nodes are distributed over a wide range, leading to steadily improved MAE results. 
This section aims to show that considering ranking fairness in SPL can obtain more reasonable results.

Tab.~\ref{headpose_table} shows the comparison results of SP-DRFs and SPUDRFs with recent head pose estimation approaches.  
Because SVR~\cite{drucker1997support}, RRF~\cite{liaw2002classification} and KPLS~\cite{al2012partial} are all conventional regression methods, they  have inferior performance relative to MoDRN.
SPUDRFs achieve the best performance with an MAE of $0.74$ on the BIWI dataset and $0.82$ on the BU-3DFE dataset, which is state-of-the-art performance. 
Such a significant gain in MAE $(1.13\rightarrow0.74)$ on BIWI demonstrates that considering ranking fairness in early paces when training DRFs can lead to much more reasonable results. 
The explanation is that, as illustrated in Fig.~\ref{Uncertainty_efficacy}, the learned leaf nodes of SPUDRFs are distributed over a wide range that can cover underrepresented examples.
 \vspace{-0.4cm}
 \begin{table}[h]
 	\centering
 	\caption{Quantitative comparison with state-of-the-art methods. \textbf{Left:} Comparison results on the BIWI dataset, \textbf{Right:} Comparison results on the BU-3DFE dataset.}
 	\label{headpose_table}
 	\begin{tabular}[h]{cc}
 		\small
 		\scalebox{1.0}{
 			\begin{tabular}{@{}l|c}
 				\hline
 				Method & MAE$\downarrow$\\
 				\hline
 				\hline
 				SVR~\cite{drucker1997support} & 3.14 \\
 				RRF~\cite{liaw2002classification} & 3.06 \\
 				KPLS~\cite{al2012partial} & 2.88 \\
 				SAE~\cite{hinton2006reducing} & 1.94 \\
 				MoDRN~\cite{huang2018mixture} & 1.62 \\
 				DRFs~\cite{shen_deep_2018} & 1.33\footnotemark[1]\\
 				SP-DRFs & 1.13 \\
 				SPUDRFs & \textbf{0.74} \\
 				\hline
 			\end{tabular}
 		}
 		&
 		\small
 		\scalebox{1.0}{
 			\begin{tabular}{@{}l|c}
 				\hline
 				Method & MAE$\downarrow$\\
 				\hline
 				\hline
 				SVR~\cite{drucker1997support} & 4.21 \\
 				SAE~\cite{hinton2006reducing} & 4.14  \\
 				KPLS~\cite{al2012partial} & 4.12 \\
 				RRF~\cite{liaw2002classification} & 4.09 \\
 				MoDRN~\cite{huang2018mixture} & 3.86 \\
 				DRFs~\cite{shen_deep_2018} &  0.99 \\
 				SP-DRFs & 0.89 \\
 				SPUDRFs & \textbf{0.82} \\
 				\hline
 			\end{tabular}
 		} \\
 	\end{tabular}
 \end{table}
\footnotetext[1]{The reported results are better than our previous work~\cite{pan2020self} because we used floating point labels in our current experiment but integral labels in~\cite{pan2020self}.}

\noindent\textbf{Results on Gaze Estimation.}
For gaze estimation, the accuracy comparisons between our proposed SP-DRFs, SPUDRFs and other baselines on the MPIIGaze dataset are shown in Tab.~\ref{MPII_table}. 
Both the MAE and standard deviation across all persons are reported. 
Note that the standard deviation of RT-GENE~\cite{fischer2018rt}, Pict-Gaze~\cite{park2018deep}, and Ordinal Loss~\cite{guo2021order} are not reported because the original studies do not provide this information. 
As shown in Tab.~\ref{MPII_table}, we observe that SPUDRFs outperform all baseline methods in MAE, confirming the effectiveness of our proposed method.
Because RT-GENE~\cite{fischer2018rt} directly maps the features extracted by VGG-16 to gaze through multiple FC layers, it has inferior performance relative to DRFs $(4.62\rightarrow4.80)$. 
For a fair comparison, we chose RT-GENE with $1$ model.
Further, because SPUDRFs method learn DRFs in a self-paced manner and take into account ranking fairness, it has a further gain in MAE over DRFs $(4.62\rightarrow4.45)$. 
Pict-Gaze~\cite{park2018deep} regresses an input image to an intermediate pictorial representation and then regresses the representation to the gaze direction.
Ordinal Loss~\cite{guo2021order} utilizes ordinal loss with order regularization to solve the regression problem.
The two methods mentioned above do not take into consideration the differences amongst samples; they train all examples simultaneously and thus have inferior MAEs. 
We observe that the MAE of Pict-Gaze~\cite{park2018deep} and Ordinal Loss~\cite{guo2021order} are $4.56$ and $4.49$ respectively, while ours is $4.45$.
It's noteworthy that SP-DRFs, when compared with DRFs, only promotes MAE slightly.
This is probably due to the obvious distribution difference between training data and test data in the leave-one-out setting.

\begin{table}[h]
	\centering
	\caption{Quantitative comparison with state-of-the-art methods on the MPIIGaze dataset.}
	\label{MPII_table}
	\begin{tabular}[h]{cc}
		\small
		\scalebox{1.0}{
			\begin{tabular}{@{}l|c}
				\hline
				Method & MAE$\downarrow$\\
				\hline
				\hline
				MPIIGaze~\cite{zhang2015appearance} & 6.30$\pm 1.80$ \\
				iTracker~\cite{krafka2016eye} & 6.20$\pm 0.85$ \\
				GazeNet+~\cite{zhang2017mpiigaze} & 5.40$\pm 0.67$\\
				MeNets~\cite{xiong2019mixed} & 4.90$\pm 0.59$\\
				RT-GENE~\cite{fischer2018rt} & 4.80$\pm -- $\\
				Pict-Gaze~\cite{park2018deep} & 4.56$\pm -- $\\
				Ordinal Loss~\cite{guo2021order} & 4.49$\pm -- $\\
				DRFs~\cite{shen_deep_2018} & 4.62$\pm 0.89 $ \\
				SP-DRFs & 4.57$\pm 0.78 $ \\
				SPUDRFs & $\textbf{4.45}\pm \textbf{0.84}$ \\
				\hline
			\end{tabular}
		}
	\end{tabular}
\end{table}

\begin{table*}
	\centering
	\caption{Quantitative Evaluation Results using Different Weighting Schemes. }
	\begin{tabular}{c|c|c|c|c|c|c|c|c|cc}
		\hline\hline
		\multirow{2}{*}{Weighting Schemes} & \multicolumn{2}{c|}{MORPH} & \multicolumn{2}{c|}{FGNET} & \multicolumn{2}{c|}{BIWI} & \multicolumn{2}{c|}{BU-3DFE} & \multicolumn{2}{c}{MPIIGaze}         \\ \cline{2-11}
		& SP-DRFs     & SPUDRFs       & SP-DRFs       & SPUDRFs      & SP-DRFs        & SPUDRFs       & SP-DRFs       & SPUDRFs       & \multicolumn{1}{c|}{SP-DRFs}   & SPUDRFs  \\ \hline\hline
		Hard         & 1.85          & \textbf{1.78}& \textbf{2.68}& 2.66          & 1.24          & 0.76         & 0.94          & 0.84         & \multicolumn{1}{c|}{4.58}          &  \textbf{4.45} \\
		Linear       &1.84 & 1.80         & 2.69         & 2.66          & 1.18          & 0.79         & 0.92          & \textbf{0.82}& \multicolumn{1}{c|}{\textbf{4.57}} & 4.46 \\
		Log          & 1.85          & 1.81         & 2.68& \textbf{2.64} & \textbf{1.13} & 0.77         & \textbf{0.89} & 0.82& \multicolumn{1}{c|}{4.60}          & 4.48 \\
		Mixture      & \textbf{1.84} & 1.80         & 2.79         & 2.70          & 1.26          & \textbf{0.74}& 0.93          & 0.86         & \multicolumn{1}{c|}{4.58}          & 4.45\\ \hline\hline
	\end{tabular}
	\label{ablation_experiments}
\end{table*}

	\subsection{Different Weighting Schemes}
A potential concern for SP-DRFs and SPUDRFs is that different weighting schemes could affect the estimation performance on a variety of visual tasks.
To evaluate this, we compare SP-DRFs/SPUDRFs under different weighting schemes, including hard, mixture, and soft weighting. 	
Under all weighting schemes, $\lambda$ was set as in Sec.~\ref{setup}. 
Under the mixture weighting scheme, $\lambda'$ was set to ensure that $10\%$ selected samples adopt soft weighting for Morph \uppercase\expandafter{\romannumeral2}, and 20\% samples for other datasets.

Table.~\ref{ablation_experiments} shows the comparison results.
The performances of SP-DRFs/SPUDRFs with different weighting schemes are only slightly different.
We observe that the weights for a large proportion of examples are close to 1.
For example, under the log weighting scheme, only $2\%$ of examples have weights below $0.5$.
Under the mixture weighting scheme, the proportion of samples whose weights are 1 can be set manually.
We chose to set this proportion to be $0.8\sim0.9$ on different tasks.
The MAEs are not guaranteed to be better than other weighting schemes.

\subsection{Robust SPUDRFs}
The intuition for SPUDRFs to work better on different visual tasks is its improved robustness, \ie, emphasizing more on ``reliable" examples. 
To further promote the robustness, we propose robust SPUDRFs in Sec.~\ref{sec:robust SPUDRFs}, which enable SPUDRFs to handle labeling noise.
We added noise to labels to test the validity of robust SPUDRFs on the above datasets. 
Specifically, we chose $10\%$ samples in Morph \uppercase\expandafter{\romannumeral2}, BIWI or MPIIGaze datasets, and added Gaussian noise $\mathcal{N}\left(0, 10\right)$ to their labels. 
To control the proportion ($0\%\sim20\%$) of samples whose likelihoods are capped to be 0, \ie, the portion of samples to be excluded, we set $\epsilon$ at variant values.

\begin{figure}
	\centering
	\subfloat{
		\includegraphics[width=0.4\textwidth]{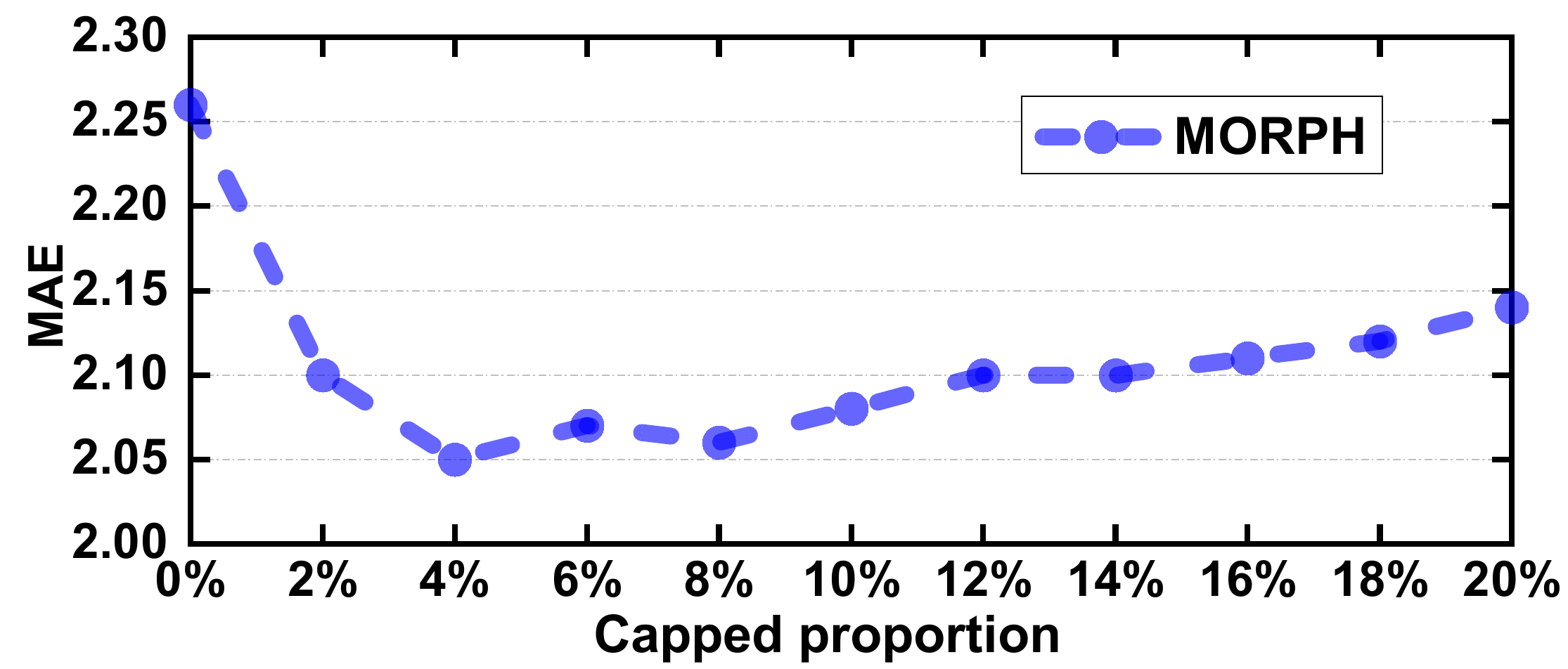}%
		\label{fig:capped_morph}}\\
	\subfloat{
		\includegraphics[width=0.4\textwidth]{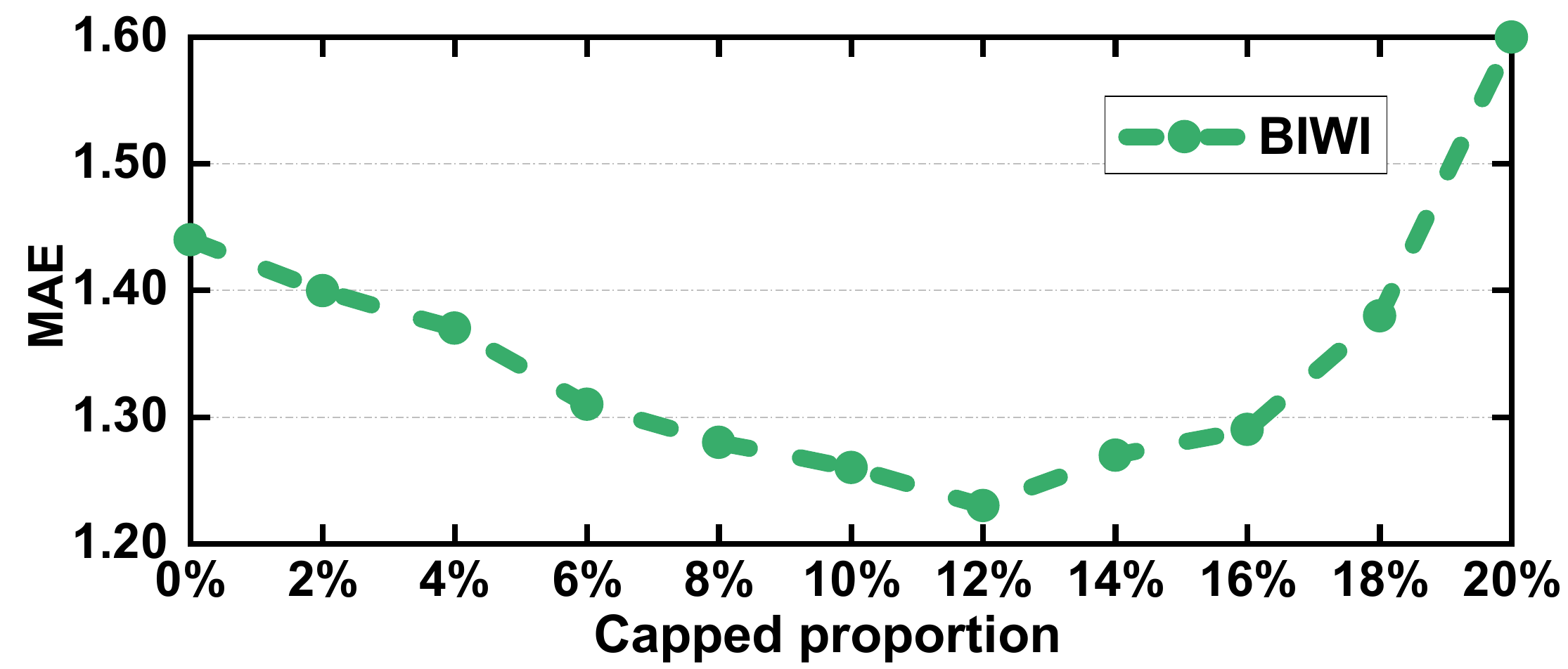}%
		\label{fig:capped_biwi}}\\
	\subfloat{
		\includegraphics[width=0.4\textwidth]{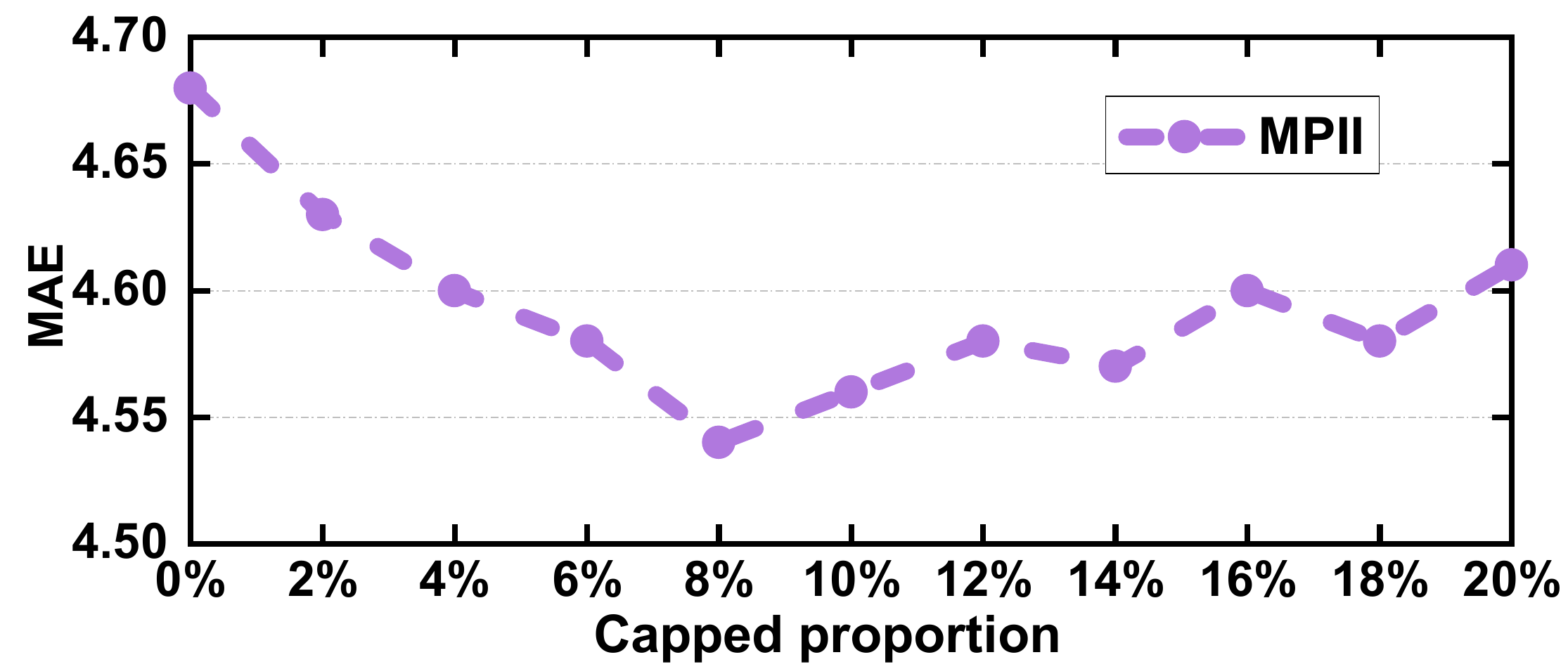}%
		\label{fig:capped_mpii}}
	\caption{The MAEs of SPUDRFs across different capped proportions evaluated on three datasets: Morph \uppercase\expandafter{\romannumeral2},  BIWI, and MPIIGaze.}
	\label{capped_experiments}
\end{figure}

Fig.~\ref{capped_experiments} shows the MAE curves of SPUDRFs across variant capped proportions. 
We observe that, when no example's likelihood is capped at 0, due to the presence of noise, the corresponding MAE ais large for each dataset. 
As the capped proportion grows, the MAE gradually decreases.
When the capped proportion changes to become $10\%$, the MAE in Fig.~\ref{capped_experiments} almost achieves minimal values, which demonstrate that robust SPUDRFs are capable of excluding noisy examples.
When the capped proportion grows continuously, the MAE changes to become large. 
One explanation is that some regular examples may be discarded when the capped proportion becomes over 10\%.

\subsection{Fairness Improvement}
\label{fairness_section}
\begin{figure*}[t]
	\centering
	\subfloat{\includegraphics[width=0.48\textwidth]{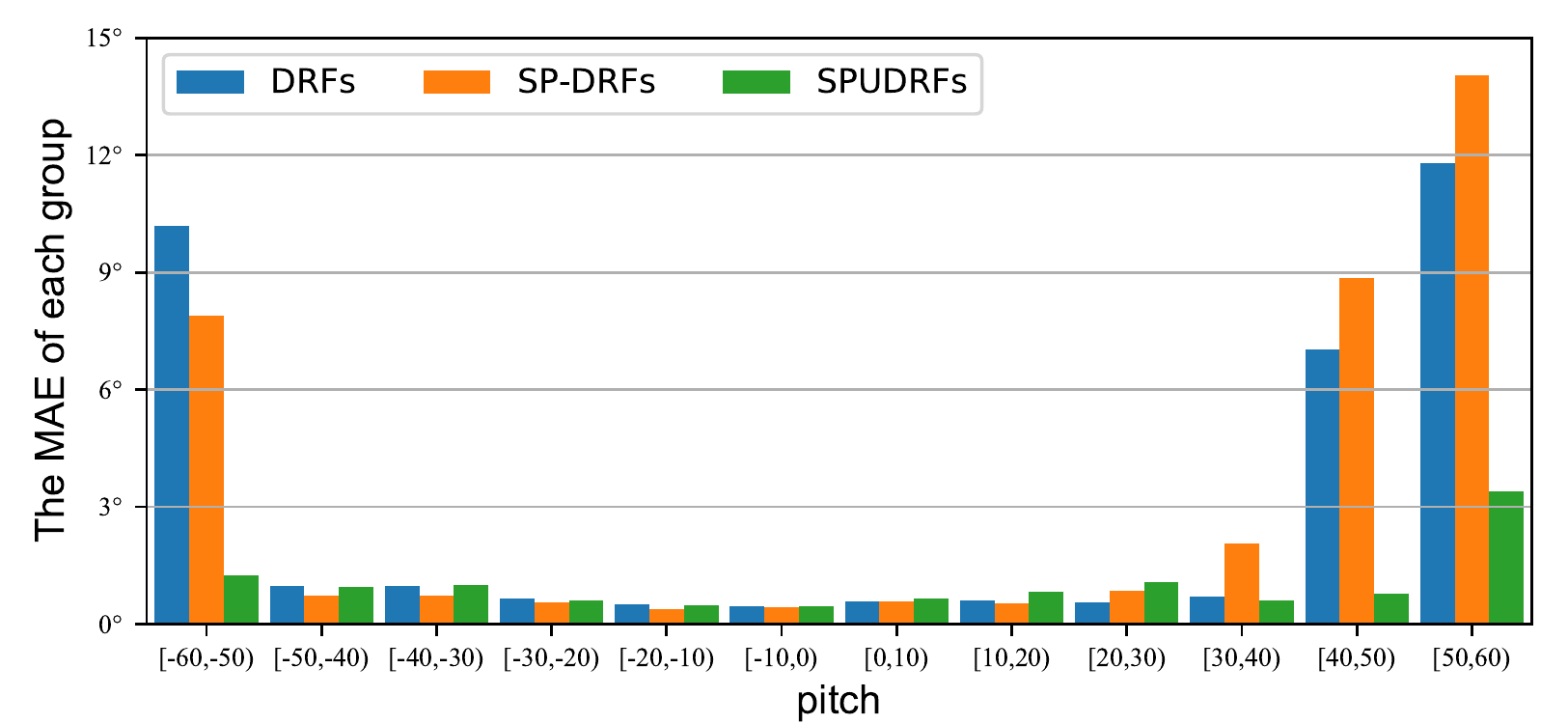}%
		\label{fairness_experiments_pitch}}
	\hfil
	\subfloat{\includegraphics[width=0.48\textwidth]{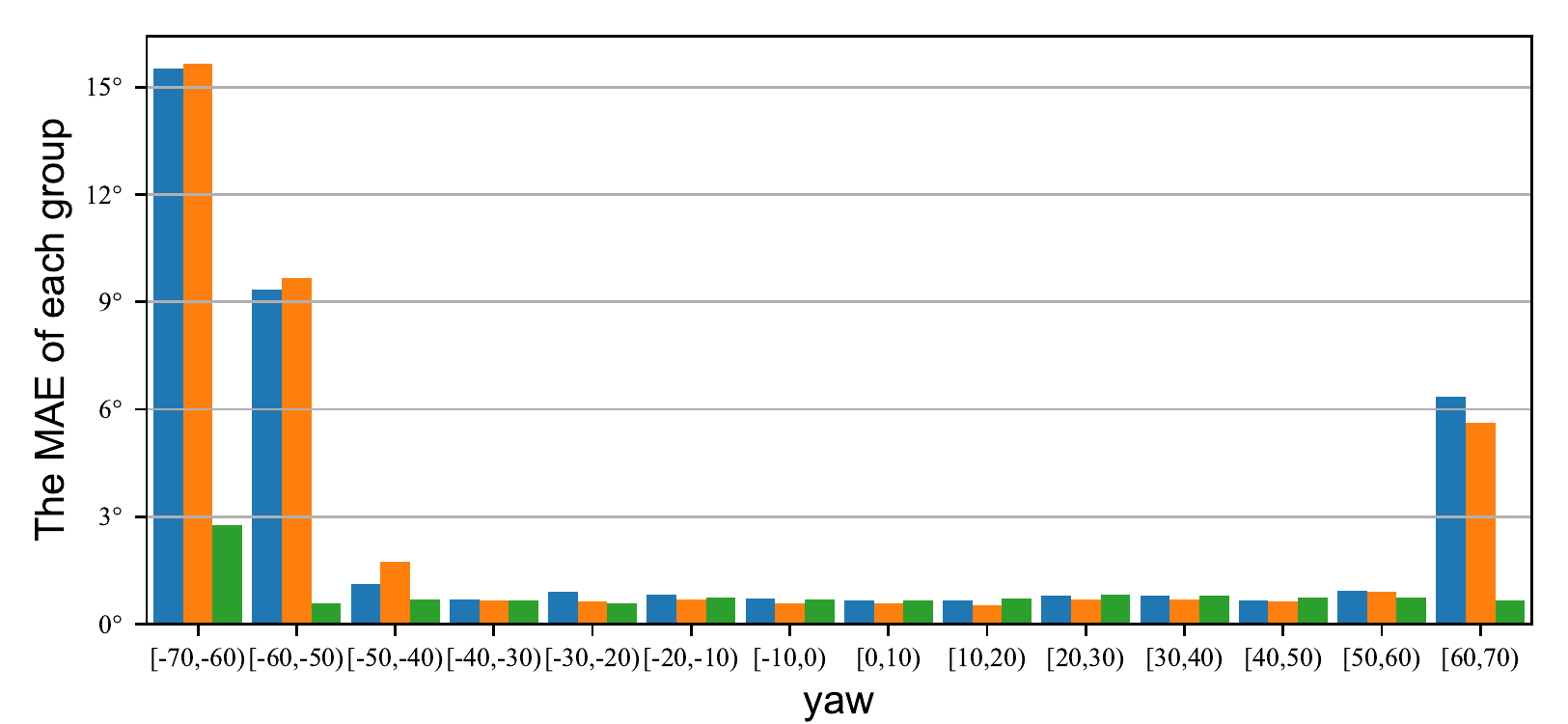}%
		\label{fairness_experiments_yaw}}
	\caption{Comparison of different methods on the BIWI dataset. In pitch and yaw directions, SP-DRFs and SPUDRFs outperform DRFs in MAE on most groups, but the MAEs of SP-DRFs are worse than DRFs on underrepresented groups.}
	\label{fairness_experiments}
\end{figure*}

This section discusses how SPUDRFs improve regression fairness on different visual tasks.

Taking into account of underrepresented examples, SPUDRFs show improved accuracy for age, pose and gaze estimation, compared to SP-DRFs.
In this section, we show how SPUDRFs further improve regression fairness.
In Sec.~\ref{sec:fairness metric}, we define FAIR as a regression fairness metric.
To evaluate the regression fairness of DRFs, SP-DRFs, and SPUDRFs, we divided all the datasets mentioned above into different subsets and calculated FAIR on them.
For Morph \uppercase\expandafter{\romannumeral2}, the entire data was divided into $7$ groups, \ie, $[10, 20], [20,30], \ldots, [70,80]$. 
A similar division was conducted on the FG-NET dataset. 
For BIWI and BU-3DFE, the pitch, yaw, and roll directions were regarded as independent directions, and the interval for each regression group is $10^\circ$. 
For MPIIGaze, the pitch and yaw directions were regarded as relevant angles.
We set the group interval to be $5^\circ$ in each direction, resulting in a total of 60 groups.

Table.~\ref{fairness_rule} shows the FAIR values on all the datasets. 
The higher the FAIR is, the more fair the regression model is.
Note that SP-DRFs always have inferior FAIR relative to DRFs.
For example, SP-DRFs have a FAIR of $0.43$, while DRFs have a FAIR of $0.46$ for BIWI, and SP-DRFs have a FAIR of $0.37$ while DRFs have a FAIR of $0.42$ for FG-NET. 
The results demonstrate that SP-DRFs tend to aggravate the bias of solutions. 
On the other hand, SPUDRFs have significantly improved FAIR compared to SP-DRFs.
We observe SPUDFRs gains improvement by $5\%$ for FGNET and $27\%$ for BIWI.
This is an evidence that SPUDRFs tend to have more fair estimation results on various visual tasks.

\begin{table}[h]
	\centering
	\caption{Regression Fairness Evaluation on Different Datasets.}
	\begin{tabular}{c|c|c|c|c|c}
		\hline
		Methods         &          Morph \uppercase\expandafter{\romannumeral2}  & FGNET & BIWI    & BU-3DFE & MPIIGaze \\
		\hline \hline
		DRFs                     & 0.46  & 0.42 & 0.46   & 0.74   & 0.67                 \\
		SP-DRFs                  & 0.44  & 0.37 & 0.43   & 0.72   & 0.67                 \\
		SPUDRFs                  & \textbf{0.48}  & \textbf{0.42} & \textbf{0.70}   & \textbf{0.76}   & \textbf{0.69}                 \\ \hline
	\end{tabular}
	\label{fairness_rule}
\end{table}	

Some pose estimation results of DRFs, SP-DRFs, and SPUDRFs for BIWI are shown in Fig.~\ref{fairness_experiments}. 
The estimation accuracy for different pose groups reveals some consistent trends.
First, for most groups, SP-DRFs and SPUDRFs have superior MAEs over DRFs. 
In Sec.~\ref{Uncertainty}, we discussed that these gains in fairness are due to SPL, which can guide DDMs to achieve more reasonable solutions.
Second, SP-DRFs tend to have even worse MAEs for the underrepresented groups than DRFs, for example, $\left[30^\circ, 60^\circ\right]$ in the pitch direction and $\left[-70^\circ, -40^\circ\right]$ in the yaw direction.
This demonstrates that existing SPL methods have a fatal drawback: the ranking and selection schemes may incur seriously biased estimation results.
Third, also for underrepresented groups, SPUDRFs gain significant improvement in MAE compared to DRFs. 
For example, our gains in MAE are $87.71\%$ and $77.68\%$ from $[-60^\circ, -50^\circ]$ and $[30^\circ, 60^\circ]$ in pitch direction, $80.96\%$ and $89.52\%$ from $[-70^\circ, -40^\circ]$ and $[60^\circ, 70^\circ]$ in yaw direction.
This also serves as an evidence that our proposed self-paced method alleviates the fairness problem in existing SPL methods.

\section{Discussion}
\label{sec:discussion}
In the experiments above, we have evaluated SPUDRFs against other baseline methods on different visual tasks, such as facial age estimation, head pose estimation and gaze estimation (Sec. \ref{sec:Comparison}).
We also evaluated the SPUDRFs method under different weighting schemes, its extension with capped likelihood formulation, and its performance improvement on fairness.
On a number of tasks and datasets, SPUDRFs and SP-DRFs outperform other baseline methods.
The advantage of considering ranking fairness in SPUDRFs is most obvious for BIWI.
For Morph II and BU-3DFE, the performance improvement when considering ranking fairness in SPUDRFs is also observable.

Learning DDMs in a self-paced manner has some limitations. 
The most noticeable one is that, in a leave-one-out setting, SP-DRFs perform only slightly better than DRFs. 
We speculate that this is due to the data distribution difference between training set and test set.
Therefore, whether SPL can improve the performance of DDMs may largely depend on the distribution divergence between training data and test data.

\section{Conclusion}
\label{sec:conclusion}
This paper explored how self-paced learning guided deep discriminative models (DDMs) to obtain robust and less biased solutions on different visual tasks, \eg~facial age estimation, head pose estimation, and gaze estimation.
A novel self-paced method, which considers ranking fairness, was proposed.
Specifically, a new ranking scheme that jointly considers loss and fairness was proposed in SPL.
Such a method was combined with a typical DDM---deep regression forests (DRFs)---and led to a new model, namely deep regression forests with consideration on underrepresented examples (SPUDRFs).
In addition, SPUDRFs under different weighting schemes, their extension with capped likelihood formulation, and their performance improvement on fairness were discussed.
Extensive experiments on three well-known computer vision tasks demonstrated the efficacy of our proposed new self-paced method.
The future work will include exploring how to incorporate such a method with other DDMs.

\section*{Acknowledgment}
The authors would like to thank Jiabei Zeng of the Institute of Computing Technology, Chinese Academy on Sciences, for the valuable advice on the gaze estimation experiments.
This work is partially supported by the National Key R\&D Program of China AI2021ZD0112000,  National Natural Science Fundation of China Nos.~62171111, 61806043, 61971106 and 61872068, and the Special Science Foundation of Quzhou No.~2020D013.

\bibliographystyle{IEEEtran}
\bibliography{mybibfile}

\begin{thebibliography}{10}
\providecommand{\url}[1]{#1}
\csname url@samestyle\endcsname
\providecommand{\newblock}{\relax}
\providecommand{\bibinfo}[2]{#2}
\providecommand{\BIBentrySTDinterwordspacing}{\spaceskip=0pt\relax}
\providecommand{\BIBentryALTinterwordstretchfactor}{4}
\providecommand{\BIBentryALTinterwordspacing}{\spaceskip=\fontdimen2\font plus
\BIBentryALTinterwordstretchfactor\fontdimen3\font minus
  \fontdimen4\font\relax}
\providecommand{\BIBforeignlanguage}[2]{{%
\expandafter\ifx\csname l@#1\endcsname\relax
\typeout{** WARNING: IEEEtran.bst: No hyphenation pattern has been}%
\typeout{** loaded for the language `#1'. Using the pattern for}%
\typeout{** the default language instead.}%
\else
\language=\csname l@#1\endcsname
\fi
#2}}
\providecommand{\BIBdecl}{\relax}
\BIBdecl

\bibitem{chendeepage}
S.~{Chen}, C.~{Zhang}, and M.~{Dong}, ``Deep age estimation: From
  classification to ranking,'' \emph{IEEE Transactions on Multimedia}, vol.~20,
  no.~8, pp. 2209--2222, 2018.

\bibitem{Yan2014Age}
C.~Yan, C.~Lang, W.~Tao, X.~Du, and Z.~Chen, ``Age estimation based on
  convolutional neural network,'' in \emph{Pacific Rim Conference on
  Multimedia}, 2014, pp. 211--220.

\bibitem{chen_using_2017}
S.~Chen, C.~Zhang, M.~Dong, J.~Le, and M.~Rao, ``Using {Ranking}-{CNN} for age
  estimation,'' in \emph{CVPR}, 2017, pp. 742--751.

\bibitem{huang2018mixture}
Y.~Huang, L.~Pan, Y.~Zheng, and M.~Xie, ``Mixture of deep regression networks
  for head pose estimation,'' in \emph{ICIP}.\hskip 1em plus 0.5em minus
  0.4em\relax IEEE, 2018, pp. 4093--4097.

\bibitem{rothe2018deep}
R.~Rothe, R.~Timofte, and L.~Van~Gool, ``Deep expectation of real and apparent
  age from a single image without facial landmarks,'' \emph{IJCV}, vol. 126,
  no. 2-4, pp. 144--157, 2018.

\bibitem{huang_soft-margin_2017}
D.~Huang, L.~Han, and F.~De~la Torre, ``Soft-margin mixture of regressions,''
  in \emph{CVPR}, 2017, pp. 4058--4066.

\bibitem{ruiz2018fine}
N.~Ruiz, E.~Chong, and J.~M. Rehg, ``Fine-grained head pose estimation without
  keypoints,'' in \emph{CVPRW}, 2018, pp. 2074--2083.

\bibitem{gao_age_2018}
B.-B. Gao, H.-Y. Zhou, J.~Wu, and X.~Geng, ``Age estimation using expectation
  of label distribution learning,'' in \emph{IJCAI}, 2018, pp. 712--718.

\bibitem{parkhi2015deep}
O.~M. Parkhi, A.~Vedaldi, and A.~Zisserman, ``Deep face recognition,'' 2015.

\bibitem{niu_ordinal_2016}
Z.~Niu, M.~Zhou, L.~Wang, X.~Gao, and G.~Hua, ``Ordinal regression with
  multiple output {CNN} for age estimation,'' in \emph{CVPR}, 2016, pp.
  4920--4928.

\bibitem{agustsson2017anchored}
E.~Agustsson, R.~Timofte, and L.~Van~Gool, ``Anchored regression networks
  applied to age estimation and super resolution,'' in \emph{ICCV}, 2017, pp.
  1643--1652.

\bibitem{cui2019class}
Y.~Cui, M.~Jia, T.-Y. Lin, Y.~Song, and S.~Belongie, ``Class-balanced loss
  based on effective number of samples,'' in \emph{CVPR}, 2019, pp. 9268--9277.

\bibitem{khan2019striking}
S.~Khan, M.~Hayat, S.~W. Zamir, J.~Shen, and L.~Shao, ``Striking the right
  balance with uncertainty,'' in \emph{CVPR}, 2019, pp. 103--112.

\bibitem{ren2018learning}
M.~Ren, W.~Zeng, B.~Yang, and R.~Urtasun, ``Learning to reweight examples for
  robust deep learning,'' \emph{arXiv preprint arXiv:1803.09050}, 2018.

\bibitem{Kumar2010Self}
M.~P. Kumar, B.~Packer, and D.~Koller, ``Self-paced learning for latent
  variable models,'' in \emph{NeurIPS}, 2010, pp. 1189--1197.

\bibitem{jiang2015self}
L.~Jiang, D.~Meng, Q.~Zhao, S.~Shan, and A.~G. Hauptmann, ``Self-paced
  curriculum learning,'' in \emph{AAAI}, 2015, pp. 2694--2700.

\bibitem{ma2017self}
F.~Ma, D.~Meng, Q.~Xie, Z.~Li, and X.~Dong, ``Self-paced co-training,'' in
  \emph{ICML}, 2017, pp. 2275--2284.

\bibitem{ijcai2017}
H.~Li and M.~Gong, ``Self-paced convolutional neural networks,'' in
  \emph{IJCAI}, 2017, pp. 2110--2116.

\bibitem{ricanek2006morph}
K.~Ricanek and T.~Tesafaye, ``Morph: A longitudinal image database of normal
  adult age-progression,'' in \emph{FG}.\hskip 1em plus 0.5em minus 0.4em\relax
  IEEE, 2006, pp. 341--345.

\bibitem{panis2016overview}
G.~Panis, A.~Lanitis, N.~Tsapatsoulis, and T.~F. Cootes, ``Overview of research
  on facial ageing using the fg-net ageing database,'' \emph{Iet Biometrics},
  vol.~5, no.~2, pp. 37--46, 2016.

\bibitem{fanelli2013random}
G.~Fanelli, M.~Dantone, J.~Gall, A.~Fossati, and L.~Van~Gool, ``Random forests
  for real time 3d face analysis,'' \emph{IJCV}, vol. 101, no.~3, pp. 437--458,
  2013.

\bibitem{pan2016mixture}
L.~Pan, J.~M. Saragih, and W.-S. Chu, ``Mixture of grouped regressors and its
  application to visual mapping,'' \emph{Pattern Recognition}, vol.~53, pp.
  184--194, 2016.

\bibitem{zhang2015appearance}
X.~Zhang, Y.~Sugano, M.~Fritz, and A.~Bulling, ``Appearance-based gaze
  estimation in the wild,'' in \emph{CVPR}, 2015, pp. 4511--4520.

\bibitem{pan2020self}
L.~Pan, S.~Ai, Y.~Ren, and Z.~Xu, ``Self-paced deep regression forests with
  consideration on underrepresented examples,'' in \emph{ECCV}, 2020.

\bibitem{meng_theoretical_2017}
D.~Meng, Q.~Zhao, and L.~Jiang, ``A theoretical understanding of self-paced
  learning,'' \emph{Information Sciences}, vol. 414, pp. 319--328, 2017.

\bibitem{Huang2021DSMVC}
Z.~Huang, Y.~Ren, X.~Pu, L.~Pan, D.~Yao, and G.~Yu, ``Dual self-paced
  multi-view clustering,'' \emph{Neural Networks}, vol. 140, pp. 184--192,
  2021.

\bibitem{Huang2021NSMVC}
Z.~Huang, Y.~Ren, X.~Pu, and L.~He, ``Non-linear fusion for self-paced
  multi-view clustering,'' in \emph{ACM MM}, 2021.

\bibitem{Guo2019adaptive}
X.~Guo, X.~Liu, E.~Zhu, X.~Zhu, M.~Li, X.~Xu, and J.~Yin, ``Adaptive self-paced
  deep clustering with data augmentation,'' \emph{IEEE Transactions on
  Knowledge and Data Engineering}, vol.~32, no.~9, pp. 1680--1693, 2019.

\bibitem{ren2018self}
Y.~Ren, X.~Yan, Z.~Hu, and Z.~Xu, ``Self-paced multi-task multi-view
  capped-norm clustering,'' in \emph{International Conference on Neural
  Information Processing}.\hskip 1em plus 0.5em minus 0.4em\relax Springer,
  2018, pp. 205--217.

\bibitem{han2017self}
L.~Han, D.~Zhang, D.~Huang, X.~Chang, J.~Ren, S.~Luo, and J.~Han, ``Self-paced
  mixture of regressions.'' in \emph{IJCAI}, 2017, pp. 1816--1822.

\bibitem{yang2019self}
J.~Yang, X.~Wu, J.~Liang, X.~Sun, M.-M. Cheng, P.~L. Rosin, and L.~Wang,
  ``Self-paced balance learning for clinical skin disease recognition,''
  \emph{IEEE transactions on neural networks and learning systems}, 2019.

\bibitem{jiang2014self}
L.~Jiang, D.~Meng, S.-I. Yu, Z.~Lan, S.~Shan, and A.~Hauptmann, ``Self-paced
  learning with diversity,'' in \emph{NeurIPS}, 2014, pp. 2078--2086.

\bibitem{shen_deep_2018}
W.~Shen, Y.~Guo, Y.~Wang, K.~Zhao, B.~Wang, and A.~Yuille, ``Deep regression
  forests for age estimation,'' in \emph{CVPR}, 2018, pp. 2304--2313.

\bibitem{li2019bridgenet}
W.~Li, J.~Lu, J.~Feng, C.~Xu, J.~Zhou, and Q.~Tian, ``Bridgenet: A
  continuity-aware probabilistic network for age estimation,'' in \emph{CVPR},
  2019, pp. 1145--1154.

\bibitem{deng2021pml}
Z.~Deng, H.~Liu, Y.~Wang, C.~Wang, Z.~Yu, and X.~Sun, ``Pml: Progressive margin
  loss for long-tailed age classification,'' in \emph{CVPR}, 2021, pp.
  10\,503--10\,512.

\bibitem{riegler2013hough}
G.~Riegler, D.~Ferstl, M.~R{\"u}ther, and H.~Bischof, ``Hough networks for head
  pose estimation and facial feature localization,'' \emph{Journal of Computer
  Vision}, vol. 101, no.~3, pp. 437--458, 2013.

\bibitem{wang2019deep}
Y.~Wang, W.~Liang, J.~Shen, Y.~Jia, and L.-F. Yu, ``A deep coarse-to-fine
  network for head pose estimation from synthetic data,'' \emph{Pattern
  Recognition}, vol.~94, pp. 196--206, 2019.

\bibitem{kuhnke2019deep}
F.~Kuhnke and J.~Ostermann, ``Deep head pose estimation using synthetic images
  and partial adversarial domain adaption for continuous label spaces,'' in
  \emph{ICCV}, 2019, pp. 10\,164--10\,173.

\bibitem{zhang2017s}
X.~Zhang, Y.~Sugano, M.~Fritz, and A.~Bulling, ``It's written all over your
  face: Full-face appearance-based gaze estimation,'' in \emph{CVPRW}, 2017,
  pp. 51--60.

\bibitem{krafka2016eye}
K.~Krafka, A.~Khosla, P.~Kellnhofer, H.~Kannan, S.~Bhandarkar, W.~Matusik, and
  A.~Torralba, ``Eye tracking for everyone,'' in \emph{CVPR}, 2016, pp.
  2176--2184.

\bibitem{fischer2018rt}
T.~Fischer, H.~J. Chang, and Y.~Demiris, ``Rt-gene: Real-time eye gaze
  estimation in natural environments,'' in \emph{ECCV}, 2018, pp. 334--352.

\bibitem{park2018deep}
S.~Park, A.~Spurr, and O.~Hilliges, ``Deep pictorial gaze estimation,'' in
  \emph{ECCV}, 2018, pp. 721--738.

\bibitem{wang2019generalizing}
K.~Wang, R.~Zhao, H.~Su, and Q.~Ji, ``Generalizing eye tracking with bayesian
  adversarial learning,'' in \emph{CVPR}, 2019, pp. 11\,907--11\,916.

\bibitem{xiong2019mixed}
Y.~Xiong, H.~J. Kim, and V.~Singh, ``Mixed effects neural networks (menets)
  with applications to gaze estimation,'' in \emph{CVPR}, 2019, pp. 7743--7752.

\bibitem{biswas2021appearance}
P.~Biswas \emph{et~al.}, ``Appearance-based gaze estimation using attention and
  difference mechanism,'' in \emph{CVPRW}, 2021, pp. 3143--3152.

\bibitem{cheng2020gaze}
Y.~Cheng, X.~Zhang, F.~Lu, and Y.~Sato, ``Gaze estimation by exploring two-eye
  asymmetry,'' \emph{IEEE Transactions on Image Processing}, vol.~29, pp.
  5259--5272, 2020.

\bibitem{cheng2020coarse}
Y.~Cheng, S.~Huang, F.~Wang, C.~Qian, and F.~Lu, ``A coarse-to-fine adaptive
  network for appearance-based gaze estimation,'' in \emph{AAAI}, vol.~34,
  no.~07, 2020, pp. 10\,623--10\,630.

\bibitem{park2019few}
S.~Park, S.~D. Mello, P.~Molchanov, U.~Iqbal, O.~Hilliges, and J.~Kautz,
  ``Few-shot adaptive gaze estimation,'' in \emph{ICCV}, 2019, pp. 9368--9377.

\bibitem{yu2019improving}
Y.~Yu, G.~Liu, and J.-M. Odobez, ``Improving few-shot user-specific gaze
  adaptation via gaze redirection synthesis,'' in \emph{CVPR}, 2019, pp.
  11\,937--11\,946.

\bibitem{chen2020offset}
Z.~Chen and B.~Shi, ``Offset calibration for appearance-based gaze estimation
  via gaze decomposition,'' in \emph{WACV}, 2020, pp. 270--279.

\bibitem{huber2008entropy}
M.~F. Huber, T.~Bailey, H.~Durrant-Whyte, and U.~D. Hanebeck, ``On entropy
  approximation for gaussian mixture random vectors,'' in \emph{IEEE
  International Conference on Multisensor Fusion and Integration for
  Intelligent Systems}, 2008, pp. 181--188.

\bibitem{jiang2014easy}
L.~Jiang, D.~Meng, T.~Mitamura, and A.~G. Hauptmann, ``Easy samples first:
  Self-paced reranking for zero-example multimedia search,'' in \emph{ACM MM},
  2014, pp. 547--556.

\bibitem{fitzsimons2019general}
J.~Fitzsimons, A.~Al~Ali, M.~Osborne, and S.~Roberts, ``A general framework for
  fair regression,'' \emph{Entropy}, vol.~21, no.~8, pp. 741--753, 2019.

\bibitem{agarwal2019fair}
A.~Agarwal, M.~Dud{\'\i}k, and Z.~S. Wu, ``Fair regression: Quantitative
  definitions and reduction-based algorithms,'' in \emph{ICML}, 2019, pp.
  120--129.

\bibitem{berk2021fairness}
R.~Berk, H.~Heidari, S.~Jabbari, M.~Kearns, and A.~Roth, ``Fairness in criminal
  justice risk assessments: The state of the art,'' \emph{Sociological Methods
  \& Research}, vol.~50, no.~1, pp. 3--44, 2021.

\bibitem{komiyama2018nonconvex}
J.~Komiyama, A.~Takeda, J.~Honda, and H.~Shimao, ``Nonconvex optimization for
  regression with fairness constraints,'' in \emph{ICML}, 2018, pp. 2737--2746.

\bibitem{zafar2017fairness}
M.~B. Zafar, I.~Valera, M.~G. Rogriguez, and K.~P. Gummadi, ``Fairness
  constraints: Mechanisms for fair classification,'' in \emph{Artificial
  Intelligence and Statistics}.\hskip 1em plus 0.5em minus 0.4em\relax PMLR,
  2017, pp. 962--970.

\bibitem{zhang_joint_2016}
K.~Zhang, Z.~Zhang, Z.~Li, and Y.~Qiao, ``Joint face detection and alignment
  using multi-task cascaded convolutional networks,'' \emph{IEEE Signal
  Processing Letters}, vol.~23, no.~10, pp. 1499--1503, 2016.

\bibitem{Simonyan2015}
K.~Simonyan and A.~Zisserman, ``Very deep convolutional networks for
  large-scale image recognition,'' \emph{arXiv 1409.1556}, 2014.

\bibitem{guo_human_2009}
G.~Guo, G.~Mu, Y.~Fu, and T.~S. Huang, ``Human age estimation using
  bio-inspired features,'' in \emph{CVPR}, 2009, pp. 112--119.

\bibitem{Huerta2014Facial}
I.~Huerta, C.~Fernández, and A.~Prati, ``Facial age estimation through the
  fusion of texture and local appearance descriptors,'' in \emph{ECCV}, 2014,
  pp. 667--681.

\bibitem{Chang2011Ordinal}
K.~Y. Chang, C.~S. Chen, and Y.~P. Hung, ``Ordinal hyperplanes ranker with cost
  sensitivities for age estimation,'' in \emph{CVPR}, 2011, pp. 585--592.

\bibitem{xin_geng_facial_2013}
X.~Geng, C.~Yin, and Z.-H. Zhou, ``Facial age estimation by learning from label
  distributions,'' \emph{IEEE Transactions on Pattern Analysis and Machine
  Intelligence}, vol.~35, no.~10, pp. 2401--2412, 2013.

\bibitem{guodong_guo_image-based_2008}
G.~Guo, Y.~Fu, C.~Dyer, and T.~Huang, ``Image-based human age estimation by
  manifold learning and locally adjusted robust regression,'' \emph{IEEE
  Transactions on Image Processing}, vol.~17, no.~7, pp. 1178--1188, 2008.

\bibitem{Yu2010Multi}
Z.~Yu and D.~Y. Yeung, ``Multi-task warped gaussian process for personalized
  age estimation,'' in \emph{CVPR}, 2010, pp. 2622--2629.

\bibitem{han_demographic_2015}
H.~Han, C.~Otto, X.~Liu, and A.~K. Jain, ``Demographic estimation from face
  images: {Human} vs. machine performance,'' \emph{IEEE Transactions on Pattern
  Analysis and Machine Intelligence}, vol.~37, no.~6, pp. 1148--1161, 2015.

\bibitem{Luu2013Contourlet}
K.~Luu, K.~Seshadri, M.~Savvides, T.~D. Bui, and C.~Y. Suen, ``Contourlet
  appearance model for facial age estimation,'' in \emph{International Joint
  Conference on Biometrics}, 2013, pp. 1--8.

\bibitem{drucker1997support}
H.~Drucker, C.~J. Burges, L.~Kaufman, A.~J. Smola, and V.~Vapnik, ``Support
  vector regression machines,'' in \emph{NeurIPS}, 1997, pp. 155--161.

\bibitem{liaw2002classification}
A.~Liaw, M.~Wiener \emph{et~al.}, ``Classification and regression by
  randomforest,'' \emph{R news}, vol.~2, no.~3, pp. 18--22, 2002.

\bibitem{al2012partial}
M.~Al~Haj, J.~Gonzalez, and L.~S. Davis, ``On partial least squares in head
  pose estimation: How to simultaneously deal with misalignment,'' in
  \emph{CVPR}, 2012, pp. 2602--2609.

\bibitem{hinton2006reducing}
G.~E. Hinton and R.~R. Salakhutdinov, ``Reducing the dimensionality of data
  with neural networks,'' \emph{science}, vol. 313, no. 5786, pp. 504--507,
  2006.

\bibitem{guo2021order}
T.~Guo, H.~Zhang, B.~Yoo, Y.~Liu, Y.~Kwak, and J.-J. Han, ``Order
  regularization on ordinal loss for head pose, age and gaze estimation,'' in
  \emph{AAAI}, vol.~35, no.~2, 2021, pp. 1496--1504.

\bibitem{zhang2017mpiigaze}
X.~Zhang, Y.~Sugano, M.~Fritz, and A.~Bulling, ``Mpiigaze: Real-world dataset
  and deep appearance-based gaze estimation,'' \emph{IEEE Transactions on
  Pattern Analysis and Machine Intelligence}, vol.~41, no.~1, pp. 162--175,
  2017.

\end{thebibliography}



\vspace{11pt}
\vspace{-33pt}
\begin{IEEEbiography}[{\includegraphics[width=1in,height=1.25in,clip,keepaspectratio]{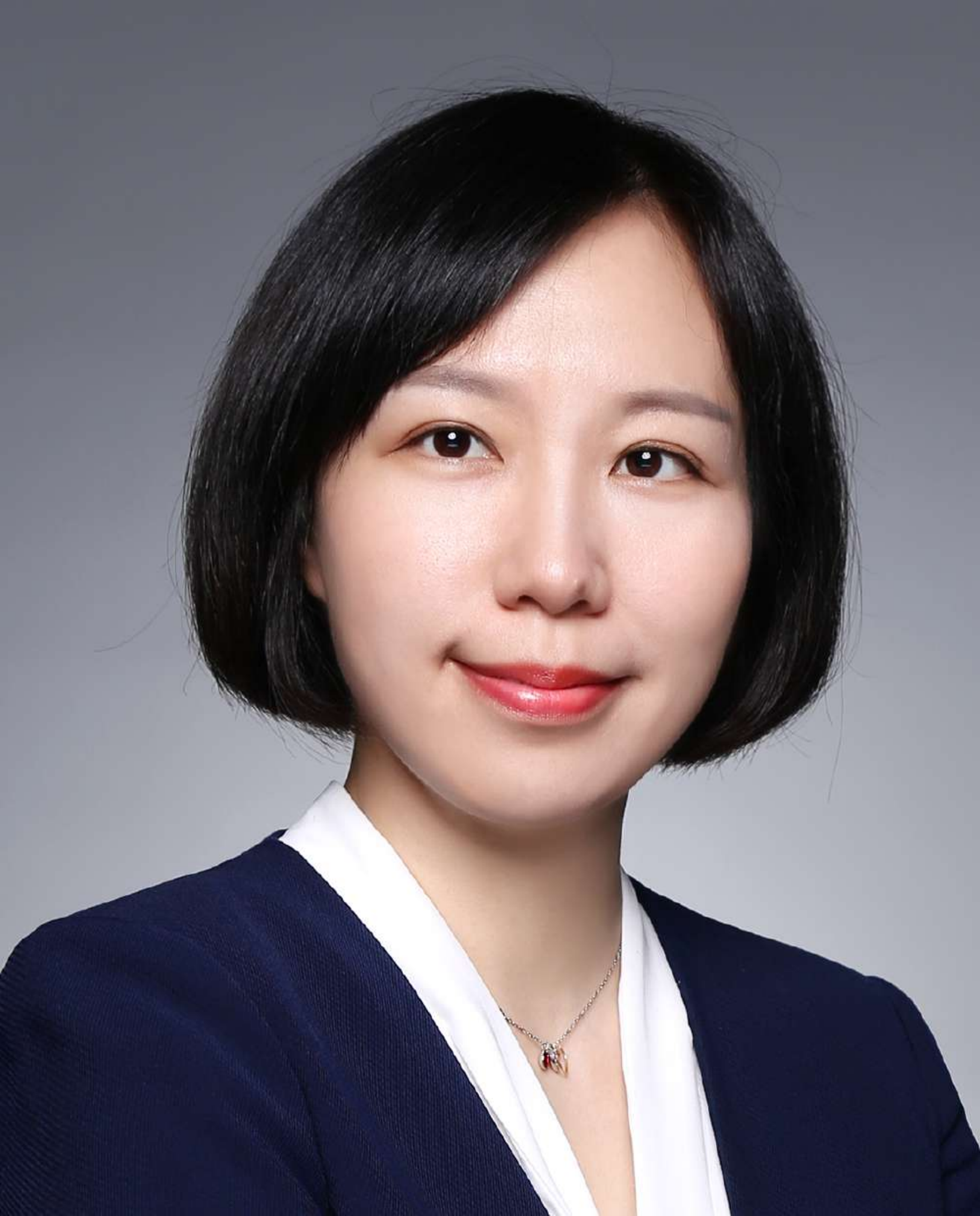}}]{Lili Pan}
	received her B.Eng.~degree in Electronic Engineering, as well as her M.Eng. and Ph.D.~degrees in Information Engineering from University of Electronic Science and Technology of China (UESTC), China, in 2004, 2007, and 2012, respectively. From 2009 to 2011, she visited the Robotics Institute of Carnegie Mellon University, USA. She joined the Department of Information Engineering, UESTC, as a lecturer in 2012. She is currently an associate professor at UESTC. Her research interests include deep discriminative models, deep generative models and interpretable machine learning.
\end{IEEEbiography}

\vspace{11pt}
\vspace{-33pt}
\begin{IEEEbiography}[{\includegraphics[width=1in,height=1.25in,clip,keepaspectratio]{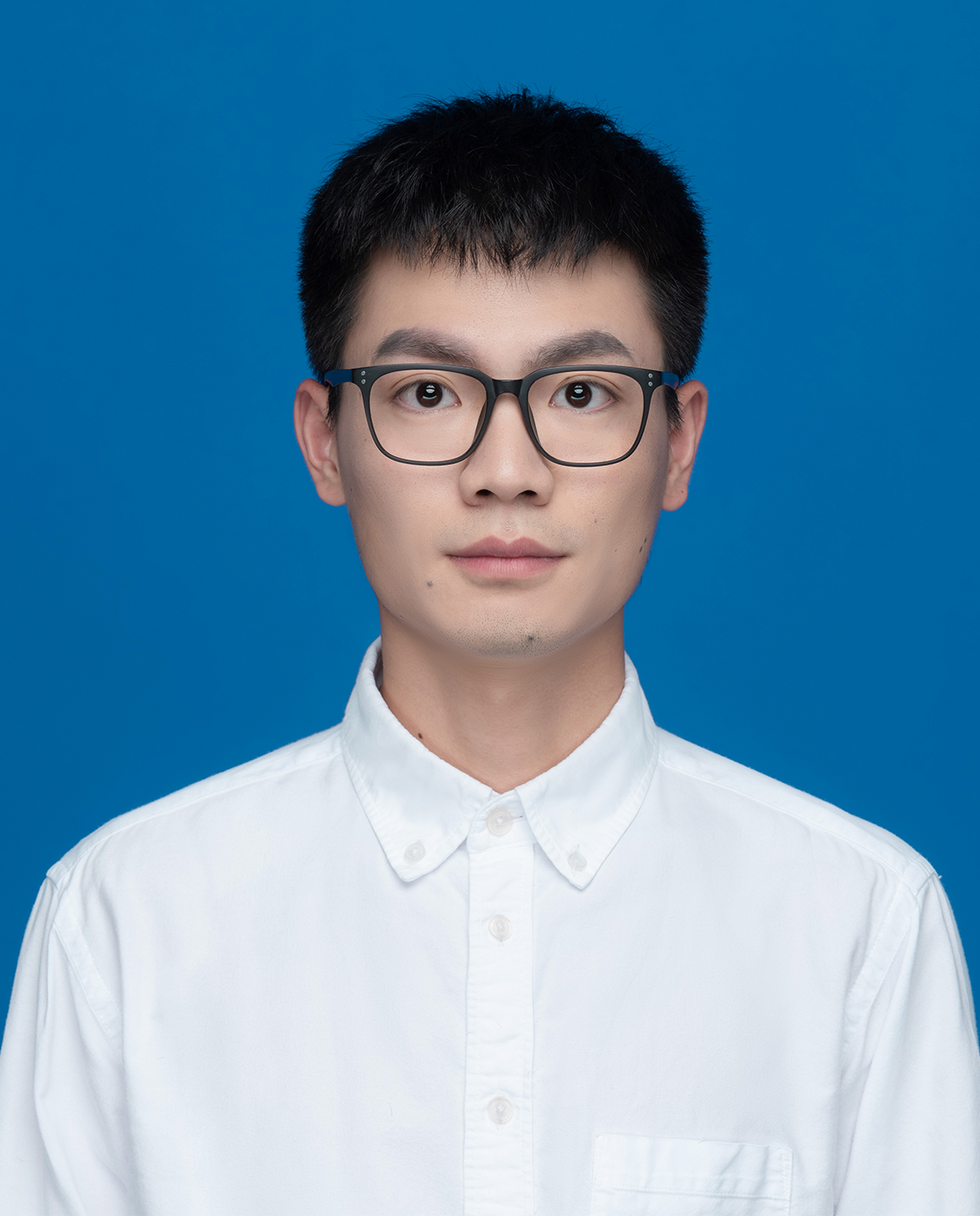}}] {Mingming Meng}
	received his B.Eng.~degree from University of Electronic Science and Technology of China (UESTC), China, in 2019. He currently is pursuing his master's degree in Information Engineering at UESTC. His research interests include deep discriminative model, fairness in deep models and generative adversarial networks. 
\end{IEEEbiography}

\vspace{11pt}
\vspace{-33pt}
\begin{IEEEbiography}[{\includegraphics[width=1in,height=1.25in,clip,keepaspectratio]{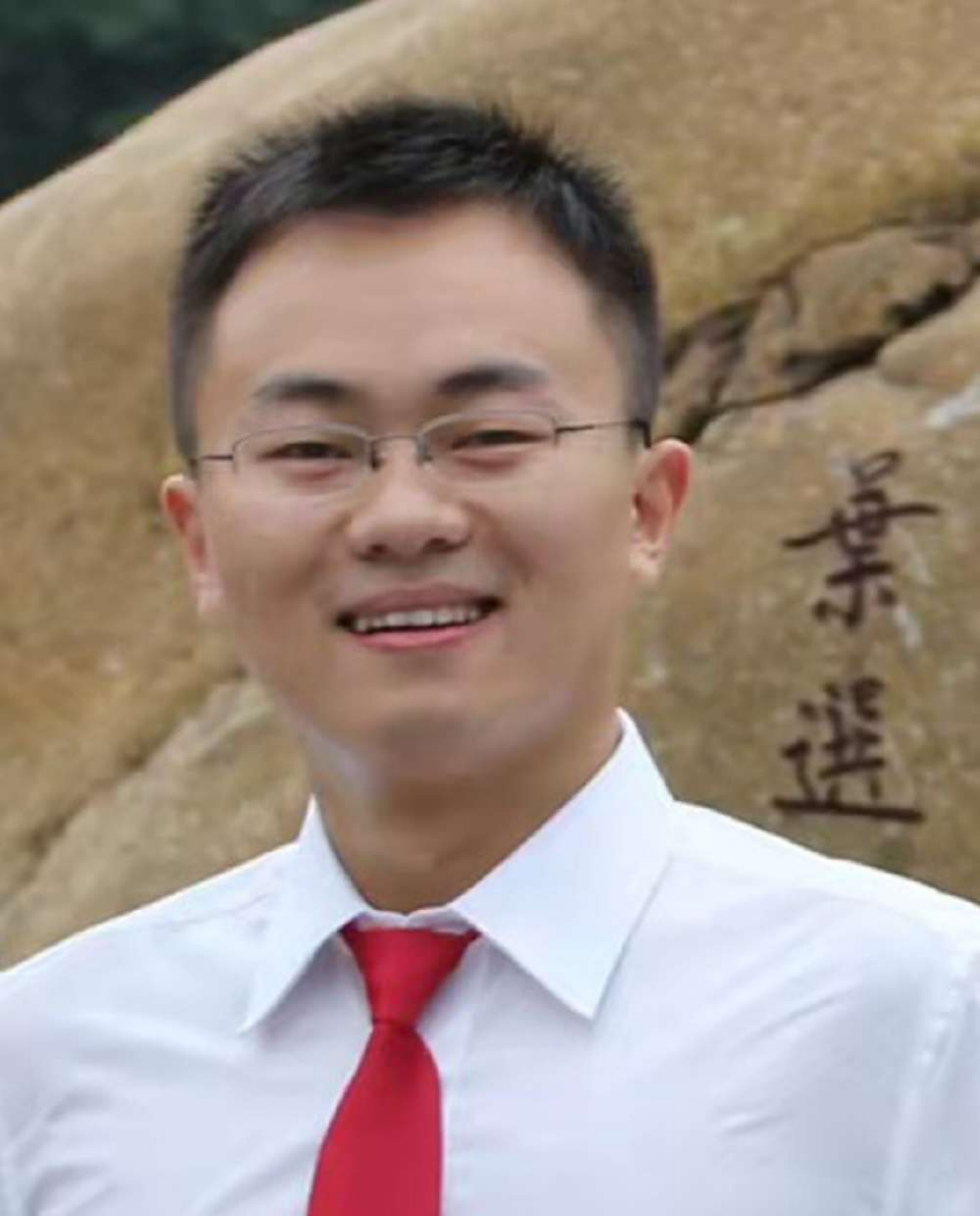}}]{Yazhou Ren}
	is currently an associate professor with the School of Computer Science and Engineering, University of Electronic Science and Technology of China, Chengdu, China. He received the B.Sc. degree in information and computation science and the Ph.D. degree in computer science from the South China University of Technology, Guangzhou, China, in 2009 and 2014, respectively. He visited the Data Mining Laboratory, George Mason University, USA, from 2012 to 2014. He has published more than 50 peer-reviewed research articles. His current research interests include clustering, self-paced learning, and multi-view learning.
\end{IEEEbiography}
\vspace{11pt}
\vspace{-33pt}
\begin{IEEEbiography}[{\includegraphics[width=1in,height=1.25in,clip,keepaspectratio]{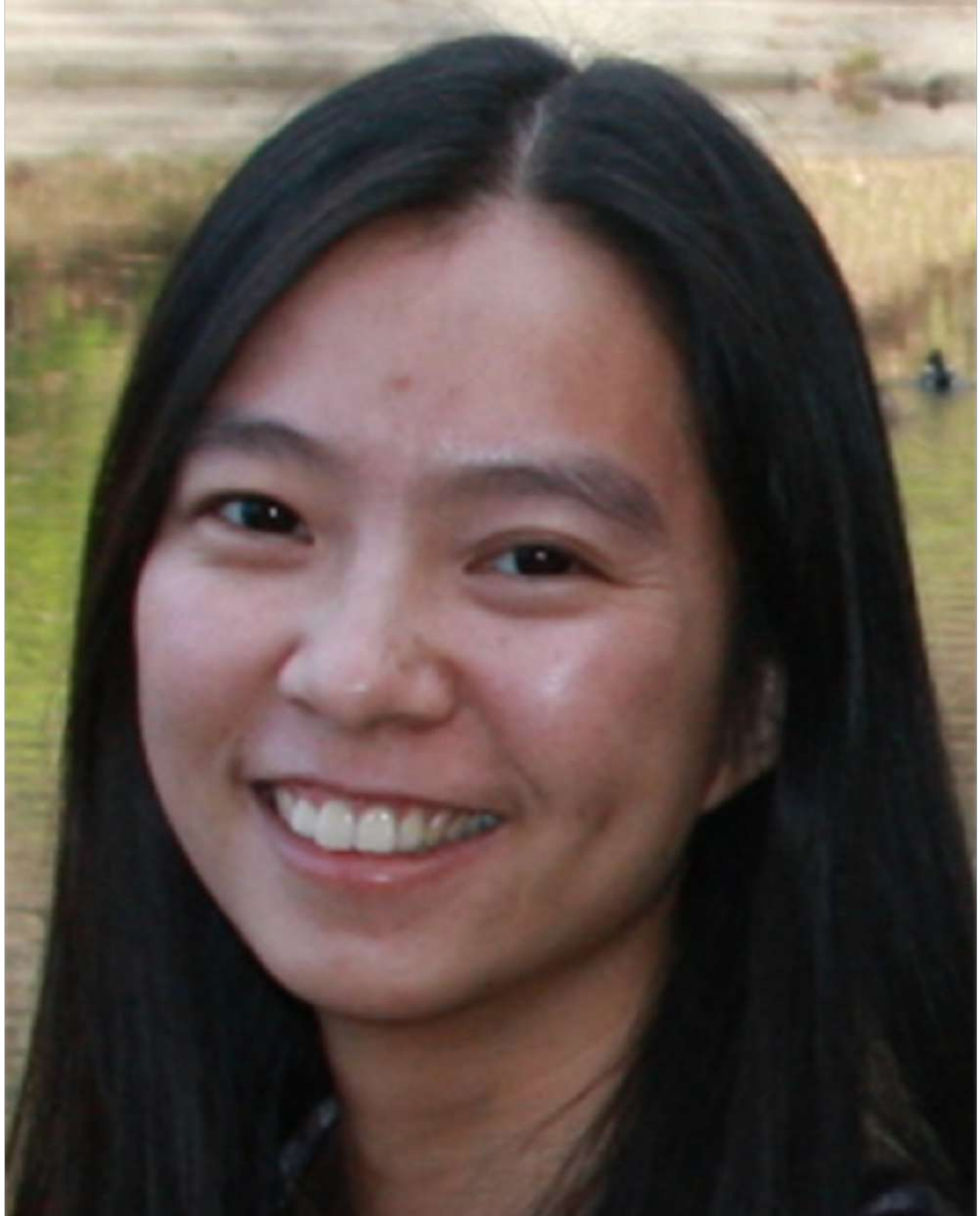}}]{Yali Zheng}
	received her PhD from Chongqing University in 2012, and joined the School of Automation Engineering, University of Electronic Science and Technology of China in 2013. From 2008 to 2011, she visited he Robotics Institute of Carnegie Mellon University, USA. Her research interests include machine learning, pattern recognition and computer vision, especially on graph matching, deep learning, and their applications in industry fields.
\end{IEEEbiography}
\vspace{11pt}
\vspace{-33pt}
\begin{IEEEbiography}[{\includegraphics[width=1in,height=1.25in,clip,keepaspectratio]{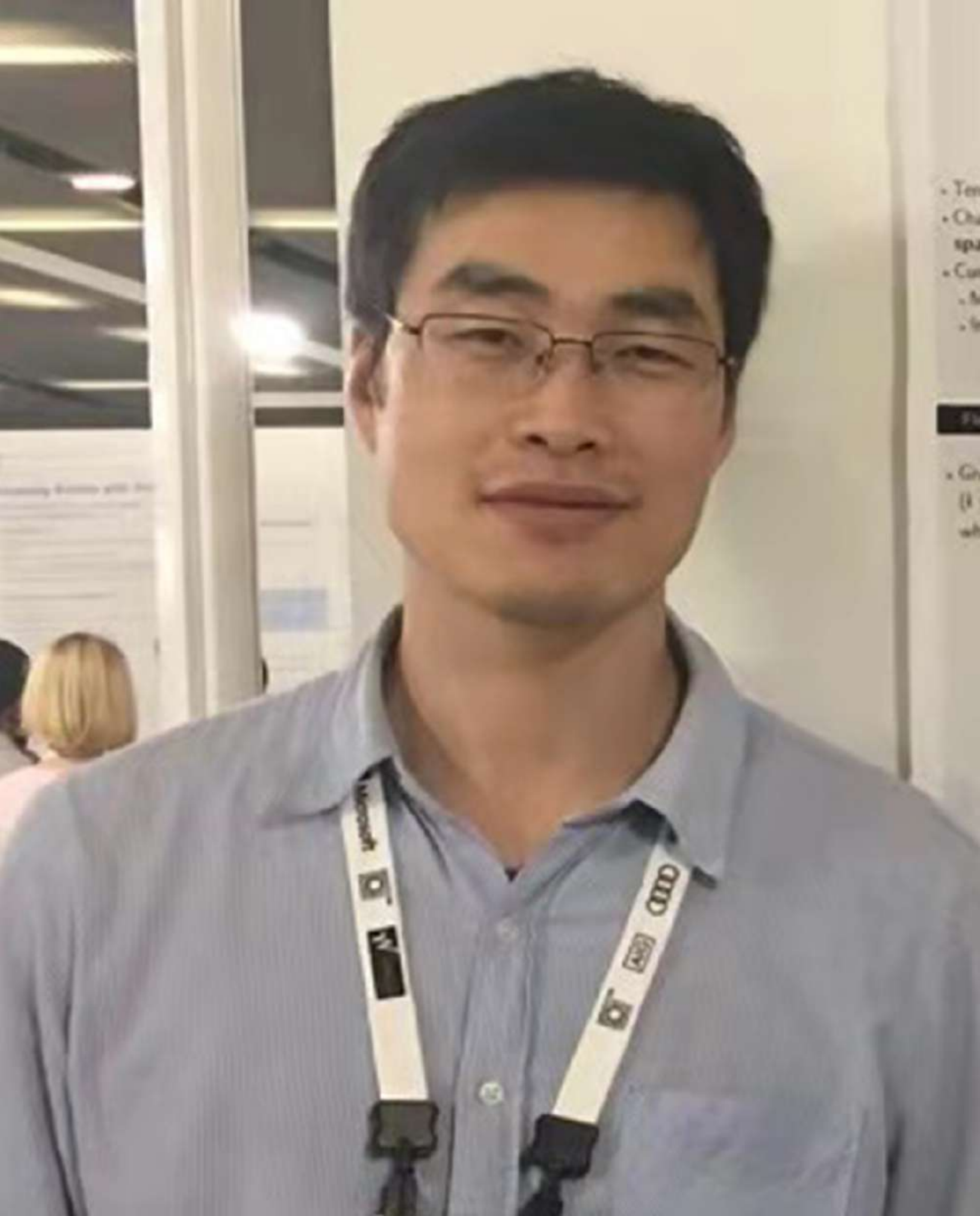}}]{Zenglin~Xu}
	received the Ph.D. degree in computer science and engineering from the Chinese University of Hong Kong. He is currently a full professor in Harbin Institute of Technology, Shenzhen, and also is affiliated with Peng Cheng Lab.   He has worked at Michigan State University, Cluster of Excellence at Saarland University, Max Planck Institute for Informatics, Purdue University, and the University of Electronic Science \& Technology of China. Dr. Xu's research interests include machine learning and its applications in computer vision, natural language processing, and health informatics. He currently serves as an associate editor to Neural Networks and Neurocomputing. He is the recipient of the outstanding student paper honorable mention of AAAI 2015, the best student paper runner up of ACML 2016, and the 2016 young researcher award from APNNS.  
\end{IEEEbiography}
\vfill

\end{document}